\documentclass{article}

\usepackage{microtype}
\usepackage{graphicx}
\usepackage{subcaption}
\usepackage{booktabs} 
\usepackage{pifont}

\usepackage{hyperref}

\usepackage[accepted]{icml2026}

\usepackage{amsmath}
\usepackage{amssymb}
\usepackage{mathtools}
\usepackage{amsthm}

\usepackage[capitalize,noabbrev]{cleveref}

\theoremstyle{plain}

\theoremstyle{definition}

\theoremstyle{remark}

\usepackage[disable,textsize=tiny]{todonotes}

\usepackage{commands}
\usepackage{xspace}
\usepackage{algorithm}
\usepackage{algorithmic}
\usepackage{tabularx}
\usepackage{siunitx}

\newcommand{\methodnamelongplain}{Entropy Coding Meets Quantization\xspace}
\newcommand{\methodname}{\mbox{EntQuant}\xspace}

\newcommand{\methodnameplain}{\mbox{EntQuant}\xspace}

\newcommand{\amethod}[1]{#1\xspace}

\newcommand{\floateight}{\texttt{Float8}\xspace}
\newcommand{\inteight}{\texttt{Int8}\xspace}
\newcommand{\floateightplain}{Float8\xspace}
\newcommand{\inteightplain}{Int8\xspace}

\newcommand{\intfour}{\texttt{Int4}\xspace}
\newcommand{\bfloat}{\texttt{BFloat16}\xspace}
\newcommand{\bfloatplain}{BFloat16\xspace}

\newcommand{\floatsixteen}{\texttt{Float16}\xspace}

\icmltitlerunning{\methodnamelongplain}

\begin{document}

\twocolumn[
  \icmltitle{Float8@2bits: Entropy Coding Enables Data-Free Model Compression}



  \icmlsetsymbol{equal}{*}

  \begin{icmlauthorlist}
  \icmlauthor{Patrick Putzky}{equal,mxm}
  \icmlauthor{Martin Genzel}{equal,mxm}
  \icmlauthor{Mattes Mollenhauer}{mxm}
  \icmlauthor{Sebastian Schulze}{mxm}
  \icmlauthor{Thomas Wollmann}{mxm}
  \icmlauthor{Stefan Dietzel}{mxm}
  \end{icmlauthorlist}
    
  \icmlaffiliation{mxm}{Merantix Momentum GmbH, Berlin, Germany}
    
  \icmlcorrespondingauthor{Patrick Putzky}{patrick.putzky@merantix-momentum.com}
  \icmlcorrespondingauthor{Martin Genzel}{martin.genzel@merantix-momentum.com}
    
  \icmlkeywords{Model Compression, Quantization, Entropy Coding, Large Language Models}

  \vskip 0.3in
]



\printAffiliationsAndNotice{\icmlEqualContribution}

\begin{abstract}
Post-training compression is currently divided into two contrasting regimes. On the one hand, fast, data-free, and model-agnostic methods (e.g., NF4 or HQQ) offer maximum accessibility but suffer from functional collapse at extreme bit-rates below 4~bits. On the other hand, techniques leveraging calibration data or extensive recovery training achieve superior fidelity but impose high computational constraints and face uncertain robustness under data distribution shifts. We introduce \methodnameplain, a framework that unites the advantages of these distinct paradigms. By matching the performance of data-dependent methods with the speed and universality of data-free techniques, \methodnameplain enables practical utility in the extreme compression regime. Our method decouples numerical precision from storage cost via entropy coding, compressing a 70B parameter model in less than 10~minutes. We demonstrate that \methodnameplain does not only achieve state-of-the-art results on standard evaluation sets and models, but also retains functional performance on more complex benchmarks with instruction-tuned models, all at modest inference overhead.
\end{abstract}

\section{Introduction}
\label{sec:intro}
\begin{figure}[t]
    \centering
    \includegraphics[width=\linewidth]{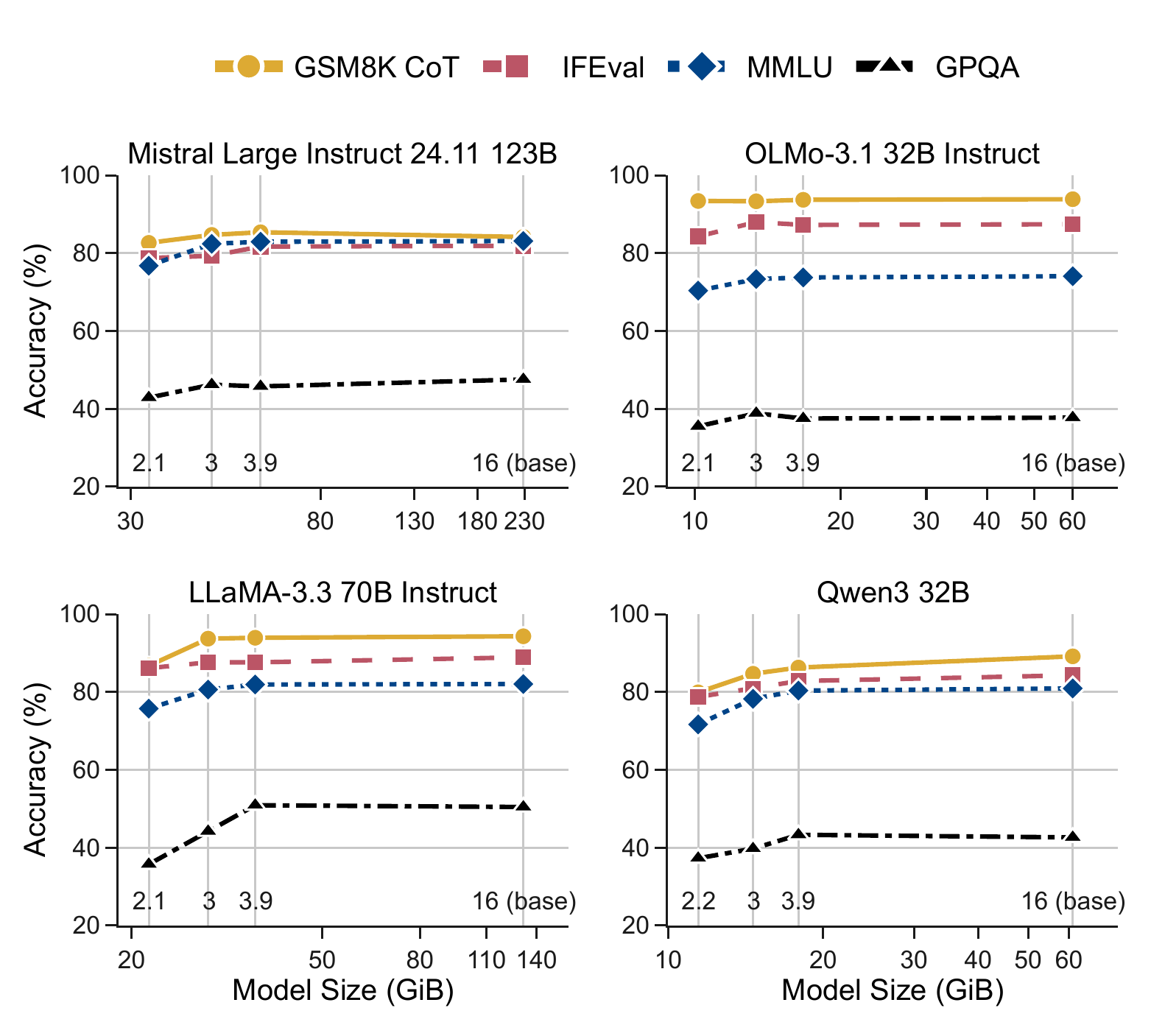}
    \caption{\methodnameplain compresses instruction-tuned models without data, performing well on several advanced benchmarks. Numbers above the size axis indicate effective bits per parameter.}
    \label{fig:figure1}
    \vspace{-.5\baselineskip}
\end{figure}

Large Language Models (LLMs) have demonstrated remarkable capabilities across a wide range of tasks \citep{touvron2023llama2openfoundation, grattafiori2024llama3herdmodels}. Yet users seeking the best model within their constraints face a trade-off between convenient API calls and self-hosting. The latter offers critical advantages in data sovereignty and latency but demands enormous memory, as open-weight models now exceed 400 billion parameters \citep{grattafiori2024llama3herdmodels,guo_deepseek-r1_2025}. Post-training quantization (PTQ) has consequently emerged as the gold standard for model weight compression~\citep{zhu_survey_2024}, reducing storage and memory-bandwidth demands while enabling faster inference \citep{frantar2023gptq,lin2024awq,frantar_marlin_2025}.

While 8-bit quantization incurs negligible degradation \citep{kurtic_give_2025, NEURIPS2022_c3ba4962, shen_efficient_2024} and \mbox{4-bit} precision remains the research frontier \citep{kurtic_give_2025}, pushing into ``extreme quantization'' below 4~bits poses significant challenges. Existing methods trade accuracy against data and compute, which \citet{nagel2019datafree} organize into four levels. The most demanding rely on recovery training or quantization-aware training (Level 3--4) \citep{egiazarian2024aqlm, tseng2024quip}; this is problematic for specialized instruction-tuned or reasoning models, whose training data is often inaccessible or legally restricted. Calibration-based PTQ (Level 2) \citep{lin2024awq, shao2024omniquant} is cheaper but still data-dependent and less accurate. At the other extreme, purely data-free methods (Level 1) such as rounding-to-nearest (\amethod{RTN}) and Half-Quadratic Quantization (\amethod{HQQ}) \citep{badri2023hqq} are fast and universal, yet collapse below 4~bits.

\begin{table}[ht]
\centering
\caption{Number of unique values in LLaMA-2 7B for different quantization levels. Averaged across layers for \methodnameplain. 2-bit \methodnameplain has more unique values than 4-bit fixed bit-width.}
\label{tab:unique_val}
\small 
\begin{tabularx}{\linewidth}{l @{\extracolsep{\fill}} S[table-format=2.2] S[table-format=2.2] S[table-format=2.2]}
\toprule
& \multicolumn{3}{c}{Quantization Levels (bits)} \\
\cmidrule{2-4}
Method & {4} & {3} & {2} \\
\midrule
Fixed bit-width & 16.00 & 8.00 & 4.00 \\
\methodnameplain $\varnothing$ & 63.89 & 49.06 & 34.61 \\
\bottomrule
\end{tabularx}
\vspace{-6pt}
\end{table}

\paragraph{Quantization Bit-Width Determines Compression Rate.}
A fundamental limitation of current quantization paradigms is the strict coupling between compression rate and representational precision. Achieving a specific storage gain requires a proportional reduction in the bit-width of the weights. For instance, quantizing from \floatsixteen to \intfour yields a compression factor of $4\times$. However, this relationship becomes a critical bottleneck in the extreme compression regime. To achieve $8\times$ compression, the standard framework dictates the use of a 2-bit representation, forcing the model to approximate complex weight distributions using only four distinct values. This lack of expressivity explains why standard quantization methods fail to preserve performance without explicitly accounting for outliers \citep{kim2024squeezellm} or resorting to extensive recovery training \citep{egiazarian2024aqlm, tseng2024quip}.

\paragraph{Decoupling Compression Rate from Bit-Width.}
Recent advancements in lossless model compression \citep{zhang_70_2025, hao2024neuzipmemoryefficienttraininginference} suggest a solution path: modern GPUs can accelerate entropy coding, leading only to modest latency penalties. This invites a pivotal question: \emph{Can we leverage entropy coding to break the strict coupling between compression rate and representational precision?}

\begin{figure*}[ht]
    \centering
    \includegraphics[width=0.95\textwidth]{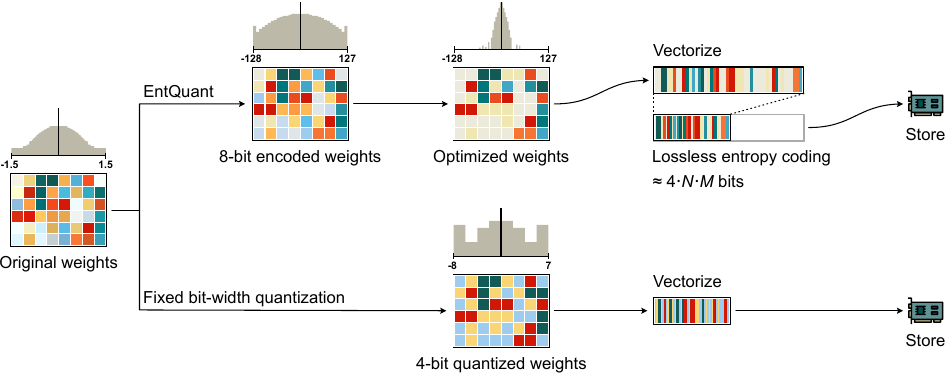}
    \caption{Illustration of 4-bit weight encoding with \methodnameplain, compared to fixed bit-width quantization. Boxes illustrate weight matrices at different representations with weight histograms above the weight matrix. Note that the number of colors and histogram bins is reduced for illustrative purposes. Weights optimized with \methodnameplain have more diverse parameters compared to fixed bit-width representations. With entropy coding, more common parameter values can be stored efficiently, see \cref{tab:unique_val}. \Cref{fig:inference_pipeline} depicts inference with \methodnameplain.}
    \label{fig:method}
\end{figure*}

In this work, we propose \methodname (\methodnamelongplain), a PTQ framework that answers this question affirmatively. Instead of forcing weights into a strict low-bit representation (e.g., 2~bits) to achieve extreme compression, we maintain a higher-precision representation (e.g., \floateight or \inteight) but optimize the weight values for \mbox{\emph{entropy}}. We then employ a GPU-optimized coding algorithm based on Asymmetric Numeral Systems (\amethod{ANS}) \citep{duda_asymmetric_2014} to losslessly compress these low-entropy weights; see \cref{fig:method} for a visual overview of \methodname.

This approach effectively decouples storage cost from representational precision. By optimizing for entropy within an 8-bit format, \methodname achieves arbitrary bit-rates per parameter, down to or even below 2~bits while running inference on high-precision kernels. Notably, the full dynamic range of the base format remains available during inference, enabling expressivity that would be impossible under rigid low-bit quantization (see \cref{tab:unique_val}). We emphasize that these benefits extend to practical deployment scenarios as shown in \cref{fig:figure1}: \methodname maintains strong performance on instruction-tuned models across challenging benchmarks, providing evidence that entropy coding offers a path beyond the limitations of fixed bit-width quantization.

\paragraph{Technical Challenges.}
We identify two key challenges to successfully employ entropy coding for model compression:

\emph{Scalable Optimization of Discrete Entropy:} Directly minimizing the entropy of a weight matrix is non-differentiable and computationally difficult. We address this by formulating a relaxed optimization objective using the $\ell_1$-norm as a differentiable proxy for entropy, enabling rapid convergence with standard gradient-based solvers.

\emph{Real-Time Decoding:} Historically, entropy coding has been viewed as a passive storage optimization, applied offline to save disk space, rather than an active component of the inference pipeline \citep{han2016deep}. Inspired by the insights from \amethod{DFloat11} \citep{zhang_70_2025}, we integrate a parallelized \amethod{ANS} decoder directly into the inference pipeline of the model, decompressing weights \emph{on-the-fly} with manageable computational overhead.

\paragraph{Contributions.} Apart from addressing the aforementioned challenges, our main contributions are as follows: 
\begin{enumerate}
    \item \emph{Calibration-Free Extreme Compression:} \methodname enables extreme PTQ (down to effective 2-bit rates) without recovery training or calibration, making it a \emph{Level-1} method in the above taxonomy and uniquely suitable for specialized (instruction-tuned or reasoning) models where data availability is a constraint.
    \item \emph{Decoupling Compression Rate from Bit-width:} \methodname introduces a method that separates the compression rate from the quantization bit-width, allowing for arbitrary compression rates while maintaining the expressiveness of \floateight or \inteight.
    \item \emph{Simplified Outlier Handling:} Unlike methods requiring explicit outlier detection \citep{NEURIPS2022_c3ba4962} or complex grouping schemes, \methodname uses only channel-wise scaling, letting the entropy optimization naturally concentrate precision where needed.
    \item \emph{High-Speed Optimization:} The \methodname compression stage is highly efficient, requiring only seconds per layer, similar to fast methods like \amethod{HQQ} \citep{badri2023hqq}, making it practical for immediate deployment.
\end{enumerate}

\paragraph{Data-Free Compression Is Essential in Practice.}
Several practical scenarios make a fully data-free approach uniquely valuable.
The most immediate is resource-constrained self-hosting, where memory is the binding constraint and calibration infrastructure is typically unavailable to end users \citep{zhang_70_2025}.
Calibration data itself is often out of reach: powerful instruction-tuned models such as LLaMA-3.3 Instruct~\citep{grattafiori2024llama3herdmodels} and Mistral Large~\citep{mistral_ai_mistral_2024} do not expose their training corpora, and regulated domains such as healthcare or finance are subject to data-protection rules (e.g., GDPR) that restrict the repurposing of sensitive data.
Even when data is accessible, calibration can degrade alignment properties of safety-tuned and reasoning models in unpredictable ways \citep{10.24963/ijcai.2025/902, wee_alignment_2025, kharinaev_investigating_2025}.
Finally, with frontier releases now arriving on a weekly cadence, multi-hour calibration pipelines have become a recurring bottleneck that a data-free method requiring less than 10 minutes eliminates entirely.

\paragraph{Software and Accessibility.}
Code is available under \url{https://github.com/merantix-momentum/entquant}.
All figures have been made colorblind safe using Paul Tol's Color Palette \citep{tol_color_2021}.

\paragraph{Conflict of Interest Disclosure.}
All authors are employed by Merantix Momentum GmbH.
The models evaluated in this work are publicly available open-weight models from third parties; none were developed by the authors' organization.
No other conflicts of interest are declared.

\section{Method}
\label{sec:method}

In this section, we introduce \methodname (\methodnamelongplain), a method designed to minimize the storage footprint of Large Foundation Models while maintaining downstream performance. Operating on a byte-level quantization format (\floateight or \inteight), we optimize the entropy of the quantized weights to approach a desired effective number of bits per parameter. \methodname integrates the decoding step directly into the forward pass with low latency overhead.

\subsection{Preliminaries}

\paragraph{Weight Quantization.}
Model compression via weight quantization reduces the precision of model weights from a high-precision format, e.g., \floatsixteen or \bfloat, to a lower-bit format such as \inteight, \floateight, or even \intfour. Given a target quantization format $\gamma$ and an $M \times N$ weight matrix $\mathbf{W}$, we denote a quantizer by $\mathbf{W}_{q} = Q_{\gamma}(\mathbf{W}, \boldsymbol\theta)$, where $\boldsymbol\theta$ is a set of parameters determining the map of $\mathbf{W}$ to its quantized form.

Dequantization brings the quantized matrix $\mathbf{W}_q$ back to a high-precision format (e.g., for inference).
This yields memory savings because only the lower-precision matrix $\mathbf{W}_q$ and parameters $\boldsymbol\theta$ need to be stored.
We denote the dequantizer by $\mathbf{\hat{W}} = Q_{\gamma}^\dagger(\mathbf{W}_q, \boldsymbol\theta)$. Thus, $Q$ and $Q^\dagger$ act as lossy encoders and decoders, respectively. Under such a quantization, a linear layer takes the form
\begin{equation*}
    \mathbf{y} =  \mathbf{x}\mathbf{\hat{W}}^\top + \mathbf{b},
\end{equation*}
where $\mathbf{x}$ and $\mathbf{y}$ are layer inputs and outputs, and $\mathbf{b}$ is a bias term. In practice, it is often unnecessary to explicitly materialize the dequantized weight matrix $\mathbf{\hat{W}}$. Instead, computation can be made more efficient by exploiting the reduced bit-width through sophisticated fused GEMM kernels like \amethod{Marlin}~\citep{frantar_marlin_2025}. Note that, for brevity, we focus on static weight quantization here although this viewpoint is compatible with dynamic activation quantization as well.

In this work, we consider symmetric weight quantization \citep{wu_integer_2020} for both \floateight and \inteight data types. The quantized weight is derived by
\begin{equation*}
     Q_{\gamma}: \mathbf{W}_{q} = \text{clamp}\left( \left\lfloor \frac{\mathbf{W}}{s} \right\rceil, -Q_{\max}, Q_{\max} \right),
\end{equation*}
where $\left\lfloor \cdot \right\rceil$ denotes rounding to the nearest representable value in data type $\gamma$ and $s$ is a scalar that adjusts weights to a desired range. A common approach is the \amethod{AbsMax Algorithm} \citep{NEURIPS2022_c3ba4962}, where $s$ is chosen as
\begin{equation}\label{eq:method:absmax}
    s = \frac{\max(|\mathbf{W}|)}{Q_{\max}},
\end{equation}
to maximize the range of values used under a given quantization. In the following, we treat $s$ as an tunable parameter. The dequantizer takes the form $Q_{\gamma}^\dagger: \mathbf{\hat{W}} = s \mathbf{W}_q$, recovering an estimate $\mathbf{\hat{W}}$ of the original weight matrix.

\paragraph{Parameter Groups.}
\label{sec:parameter_groups}
Instead of a global scaling parameter~$s$ which is used for the entire matrix, it is common practice to rescale groups of parameters $\mathbf{W}^{(g)}$ separately \citep{frantar2023gptq,lin2024awq,shao2024omniquant,badri2023hqq}. The set of all scale parameters $s^{(g)}$ is denoted by $S$. Generally, smaller group sizes yield better approximations of the original matrix but incur additional storage overhead through $|S|$. Intuitively, by optimizing the scale parameters, we can align weight distributions across groups, so that the entropy of the quantized weights can be reduced.

In \methodname, we consider channel-wise scaling, i.e., one scaling factor $s^j$ per output channel, which produces only marginal memory and inference time overhead and does not require specialized kernels.

\paragraph{Entropy Coding.}
Shannon's source coding theorem \citep{shannon_mathematical_1948} establishes that the optimal average code length for lossless compression of a sequence of i.i.d.~random variables $X$ is bounded by the entropy
\begin{equation*}\label{eq:entropy}
    H(X) := - \sum_{x \in \mathcal{X}} p(x) \log_2 p(x).
\end{equation*}
Several algorithms exist to approach this bound. Huffman coding \citep{huffman_method_1952} uses variable-length prefix-free codes but is suboptimal when symbol probabilities are not negative powers of two ($2^{-k}$) or when $H(X) < 1$. Arithmetic Coding (\amethod{AC}) resolves these inefficiencies by encoding an entire sequence into a single rational number, offering superior compression rates \citep{rissanen1976generalized, pasco1976source}. However, in its basic form, \amethod{AC} requires computationally expensive division operations and strict serial dependencies, limiting its throughput on modern GPUs. Asymmetric Numeral Systems (\amethod{ANS}) \citep{giesen_interleaved_2014} achieves the compression ratios of \amethod{AC} using only faster multiplication and bit-shift operations, which makes it significantly more efficient for massive parallelization \citep{duda_asymmetric_2014}. \amethod{ANS} has become the basis for many modern compression algorithms, such as \amethod{Zstandard} \citep{rfc8878}, and is available on NVIDIA GPUs via \amethod{nvCOMP} \citep{nvcomp} and \amethod{DietGPU} \citep{dietgpu}, and on AMD GPUs via \amethod{hipANS} \citep{hipANS}. The results of the present work are based on the \amethod{nvCOMP} package.

\subsection{\methodnameplain: \methodnamelongplain}
\label{sec:method:entquant}

The primary goal of \methodname is to minimize the entropy of (mildly) quantized weights $\mathbf{W}_q$ to achieve stronger compression. Practical entropy coding implementations rely on the empirical distribution $\hat{p}$ under an i.i.d.~assumption. For the parameters in $\mathbf{W}_q$, we assume the factorization
\begin{equation*}
    \hat{p}(\mathbf{W}_q) = \prod_{i, j} \hat{p} (\mathbf{W}_q^{(i,j)}),
\end{equation*}
where $\hat{p}(x) := \frac{1}{MN} \sum_{k=1}^{MN} \delta_x(x_k)$ represents the frequency of a value $x$ in $\mathbf{W}_q$.
Consequently, the empirical entropy (expected bits per parameter) is given by
\begin{equation} \label{eq:emp_entropy}
    \hat{H}(\mathbf{W}_q) = - \frac{1}{MN} \sum_{i, j} \log_2 \hat{p} (\mathbf{W}_q^{(i,j)}).
\end{equation}
Ideally, we would optimize $\mathbf{W}_q$ by minimizing \eqref{eq:emp_entropy} subject to an $\epsilon$-constraint on the reconstruction error:
\begin{equation*}
    \min_{\mathbf{W}_q} \hat{H}(\mathbf{W}_q) \quad \text{subject to} \quad d(\mathbf{W}, \mathbf{\hat{W}}) < \epsilon, 
\end{equation*}
where $d(\mathbf{W}, \mathbf{\hat{W}})$ measures the error of approximating $\mathbf{W}$ by~$\mathbf{\hat{W}}$. However, solving this combinatorial problem directly is computationally challenging.

We therefore relax it into a Lagrangian formulation, also referred to as rate-distortion optimization \citep{sullivan_rate-distortion_1998} or entropy-constrained optimization \citep{chou_entropy-constrained_1989}:
\begin{equation} \label{eq:optimization}
    \min_{\mathbf{W}_q} d(\mathbf{W}, \mathbf{\hat{W}}) + \lambda R(\mathbf{W}_q),
\end{equation}
where $R(\mathbf{W}_q)$ acts as a differentiable surrogate for $\hat{H}(\mathbf{W}_q)$ and the regularization parameter $\lambda > 0$ controls the compression rate. 

As reconstruction loss, we use the relative, entry-wise \mbox{$\ell_1$-loss}, which is robust to outliers:
\begin{equation*}
    d(\mathbf{X}, \mathbf{\hat{X}}) \coloneq ||\mathbf{X} - \mathbf{\hat{X}}||_1 / ||\mathbf{X}||_1, 
\end{equation*}
and as regularizer, we choose the entry-wise $\ell_1$-norm:
\begin{equation*}
    R(\mathbf{X}) \coloneq ||\mathbf{X}||_1.
\end{equation*}
We find that the $\ell_1$-norm serves as an effective and robust proxy for entropy reduction in all considered settings. A formal max-entropy bound justifying this choice is derived in \cref{app:entropy_regularization}. Additionally, $\ell_1$-regularization empirically induces a certain degree of sparsity in the resulting weights, see \cref{fig:entropy_vs_sparsity}. \cref{app:sparsity} further shows analytically that sparsity alone does not account for the observed entropy reductions, and provides a layer-wise breakdown of entropy and reconstruction error.

To solve \eqref{eq:optimization}, we initialize the weights using the \amethod{AbsMax Algorithm} from \eqref{eq:method:absmax} and optimize each layer separately using \mbox{\amethod{L-BFGS}} \citep{liu_limited_1989} in \amethod{PyTorch} \citep{Ansel_PyTorch_2_Faster_2024}. As discussed in \cref{sec:parameter_groups}, we tune only the scale parameters $S$, which leads to fast optimization even for large matrices. We use the straight-through estimator for~$Q_\gamma$ \citep{bengio2013estimatingpropagatinggradientsstochastic} to enable gradient computations through the quantization process.
As shown in \cref{fig:entropy_vs_reg_param}, the relationship between $\lambda$ and the target entropy is log-linear and largely model-independent. This strongly facilitates the choice of~$\lambda$ in practice.
Unless stated otherwise, we use \floateight as the base quantization format in \methodname; see \cref{sec:experiments:sw} for an ablation with \inteight.

\paragraph{Weight encoding.}
By solving \eqref{eq:optimization}, we obtain an optimized quantized weight matrix $\mathbf{W}_q$ and a set of scales~$S$ for each linear layer. To store this data efficiently, we treat the weights as a stream of symbols to be compressed. First, the two-dimensional matrix~$\mathbf{W}_q$ of dimensions $M \times N$ is flattened into a one-dimensional symbol sequence. Using an \amethod{ANS} coder, we compress this sequence into a compact bitstream~$\mathbf{z}$.
The final storage footprint of each layer thus consists of the compressed bitstream~$\mathbf{z}$, the quantization scales~$S$, and the metadata required by the \amethod{ANS} decoder (e.g., the symbol frequency table).

Since $|S| \ll M \cdot N$, the storage overhead of the high-precision (\bfloat) scales is negligible. Consequently, the effective compression ratio is primarily determined by the entropy of~$\mathbf{W}_q$, which we explicitly minimized in~\eqref{eq:optimization}. 
The full encoding procedure is summarized in \cref{alg:encoding}.
Notably, this constitutes a \emph{data-free} compression scheme, as only the weight matrix~$\mathbf{W}$ is required as input.

\begin{algorithm}[h]
\caption{Weight Encoding Scheme}
\label{alg:encoding}
\begin{algorithmic}[1]
    \REQUIRE Weights $\mathbf{W}$
    \ENSURE Bitstream $\mathbf{z}$, scales $S$, metadata $\mathcal{M}$
    \STATE $S^0 \leftarrow \text{AbsMax}(\mathbf{W})$ \COMMENT{Initialize with \eqref{eq:method:absmax}}
    \STATE $S^* \leftarrow \arg\min_S d(\mathbf{W}, \mathbf{\hat{W}}) + \lambda R(\mathbf{W}_q)$ \COMMENT{Solve \eqref{eq:optimization}}
    \STATE $\mathbf{W}_q \leftarrow Q_\gamma(\mathbf{W}, S^*)$ 
    \STATE $\mathbf{w} \leftarrow \text{vec}(\mathbf{W}_q)$ \COMMENT{View as 1D vector}
    \STATE $\mathbf{z} \leftarrow \text{ANS}(\mathbf{w})$ \COMMENT{Entropy encoding}
    \STATE \textbf{return} $(\mathbf{z}, S^*, \mathcal{M})$
\end{algorithmic}
\end{algorithm}

\paragraph{Inference-time Decoding.}
In standard quantization pipelines, quantized weights are typically loaded directly into GPU memory. In \methodname, we introduce an on-device decoding step that keeps weights in their highly efficient bitstream format $\mathbf{z}$ in VRAM, decompressing them \emph{on-the-fly} when required for a forward pass.
This inference procedure is formalized in \cref{alg:decoding}.

\begin{algorithm}[h]
\caption{Inference-Time Decoding Scheme}
\label{alg:decoding}
\begin{algorithmic}[1]
    \REQUIRE Input $\mathbf{x}$, bitstream $\mathbf{z}$, scales $S$, metadata $\mathcal{M}$
    \ENSURE Output $\mathbf{y}$
    \STATE $\mathbf{w} \leftarrow \text{ANS}^{-1}(\mathbf{z}, \mathcal{M})$  \COMMENT{Entropy decoding}
    \STATE $\mathbf{W_q} \leftarrow \text{view}(\mathbf{w})$ \COMMENT{View as weight matrix}
    \STATE $\mathbf{y} \leftarrow \text{QMatMul}(\mathbf{W}_q, S, \mathbf{x}) + \mathbf{b}$
    \STATE \textbf{return} $\mathbf{y}$
\end{algorithmic}
\end{algorithm}

\begin{figure}[h]
    \centering
    \includegraphics[width=\linewidth]{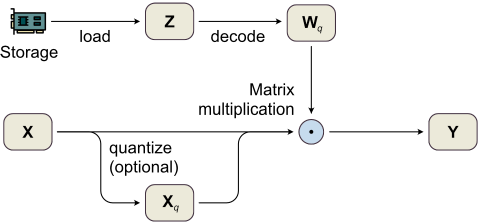}
    \caption{Visualization of \methodnameplain's inference pipeline.}
    \label{fig:inference_pipeline}
\end{figure}

While this decoding step introduces computational overhead compared to reading raw uncompressed weights, the process is designed to be highly efficient. GPU-based \amethod{ANS} decoders are highly parallelized, and for the large weight matrices of Foundation Models, they run with high hardware utilization.
See \cref{fig:method} above and \cref{fig:inference_pipeline} below for a visualization of \methodname's compression and inference pipeline, respectively, and \cref{app:implementation:entquant} for further implementation details and optimizations of \cref{alg:encoding,alg:decoding}.

\paragraph{Block-Wise Decompression Pipeline.}
In practice, we jointly compress all entropy-optimized weights of a transformer block into a single bitstream.
During inference, the model maintains one decompression buffer per device, sized to fit one transformer block in the base format (e.g., \floateight).
Before the forward pass of each block, \amethod{nvCOMP} jointly decompresses all weights into this buffer; individual layer weights are accessed via tensor views, avoiding copies (Line~2 of \cref{alg:decoding}).
After the forward pass completes, the buffer is overwritten by the next block's weights.
This block-wise scheme, following the design of \amethod{DFloat11}~\citep{zhang_70_2025}, yields an approximately 50\% inference speed-up over na\"ive layer-wise decoding.
GPU profiling (\cref{fig:nsight}) shows dense alternation between decompression and forward kernels, indicating that both stages keep the device well-utilized.

\paragraph{Memory Footprint.}
\Cref{tab:memory_footprint} reports the inference-time memory breakdown for LLaMA-2 70B at 2.1 effective bits per parameter.
The compressed weights dominate storage; all auxiliary components (scales, ANS metadata, decompression buffer) add less than 5\% overhead.
Peak GPU memory during inference is systematically analyzed in \cref{fig:efficiency_full_peak_memory}.

\begin{table}[h]
\centering
\caption{Memory footprint of \methodname for LLaMA-2 70B at 2.1 effective bits per parameter. KV cache assumes batch size~1 at the native 4096-token context with 16-bit activations.}
\label{tab:memory_footprint}
\footnotesize
\begin{tabular*}{\columnwidth}{l@{\extracolsep{\fill}}r}
\toprule
Component & Size \\
\midrule
Compressed weights & $\sim$18.8\,GiB \\
Scale parameters & $\ll 1\%$ \\
ANS metadata & negligible \\
Decompression buffer & $\sim$0.8\,GiB \\
KV cache & $\sim$1.25\,GiB \\
\bottomrule
\end{tabular*}
\end{table}

\section{Experiments}
\label{sec:experiments}

With over 480 runs, we have evaluated \methodname on 16 different open-weight LLMs, including \mbox{LLaMA-1}, \mbox{LLaMA-2} \citep{touvron2023llama2openfoundation}, \mbox{LLaMA-3.1/3.3} Base \& Instruct~\citep{grattafiori2024llama3herdmodels}, Qwen3~\citep{yang2025qwen3technicalreport}, OLMo~3.1 Instruct~\citep{olmo2025olmo3}, and Mistral Large Instruct 24.11~\citep{mistral_ai_mistral_2024}. Following standard practice, we report perplexity on C4~\citep{2020t5} and \mbox{WikiText-2}~\citep{merity2017pointer}, as well as zero-shot accuracy on eight tasks from the EleutherAI LM Evaluation Harness (LM Eval)~\citep{eval-harness}. Beyond these standard benchmarks, we assess \methodname on instruction-tuned models using more challenging evaluations including GSM8K CoT and IFEval. Implementation details on \methodname, baselines, and evaluations are provided in \cref{app:implementation}.

\subsection{\methodnameplain Outperforms Other Data-Free Methods}
\label{sec:experiments:baseline}

\begin{figure}[t]
    \centering
    \includegraphics[width=\linewidth]{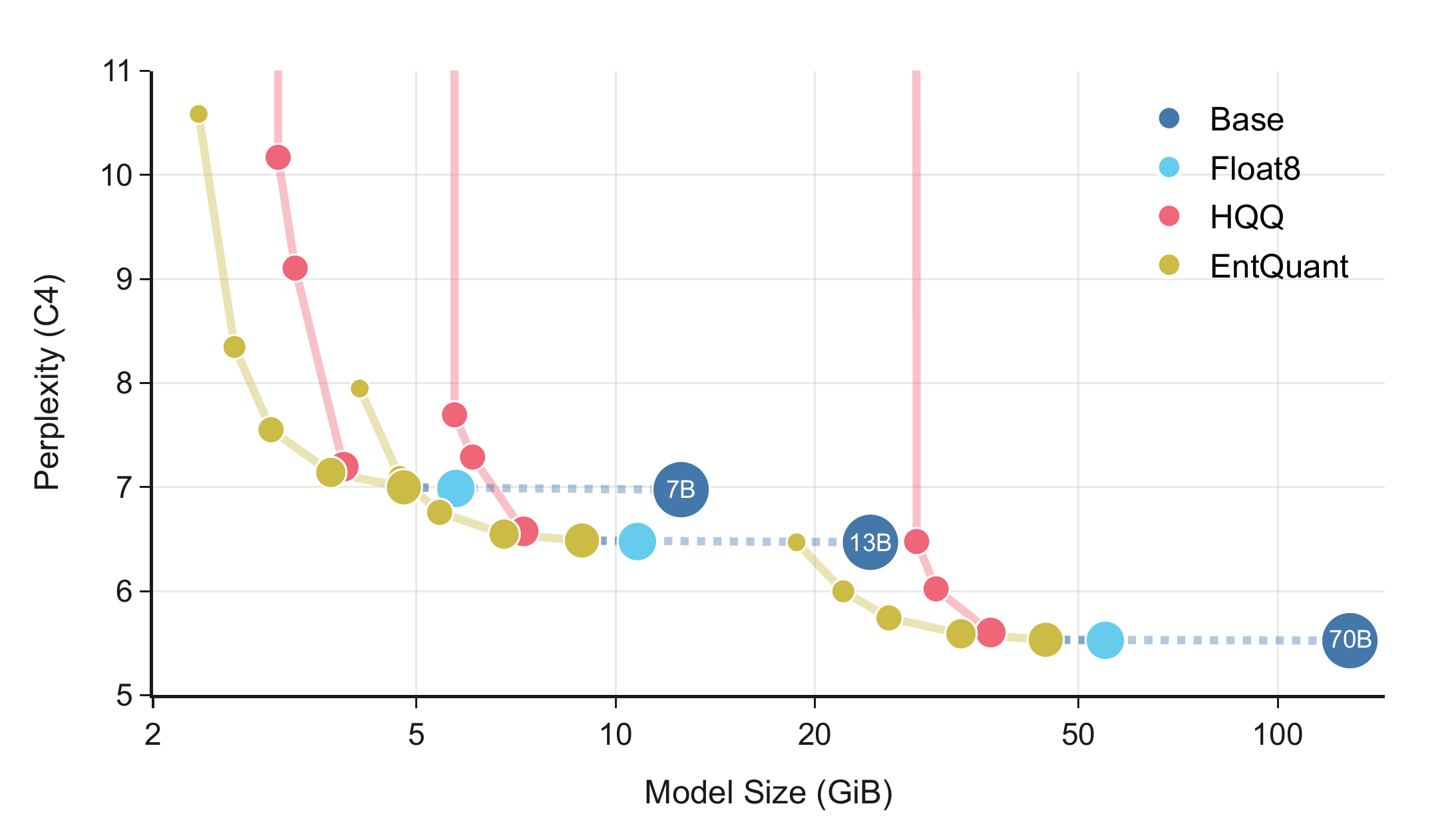}
    \caption{Memory-perplexity trade-off on C4 for LLaMA-2 7B, 13B, 70B. \methodnameplain spans a smooth Pareto front enabling fine-grained compression-performance trade-offs. Surface areas of dots are proportional to the bit-rate of each model. \floateightplain is entropy-encoded as well, leading to approximately $6.5$~bits per parameter.}
    \label{fig:memory_vs_ppl}
    \vspace{-6pt}
\end{figure}

\begin{table*}[t]
\centering
\caption{Comparison of data-free compression methods on the LLaMA base model families (example: ``1-7'' means LLaMA-1 7B). All results are generated in-house, see also \cref{app:implementation:baseline}. Best results per model and bit-rate group are in \textbf{bold}. See \cref{tab:baseline_full_llama1,tab:baseline_full_llama2,tab:baseline_full_llama31} for full results. The full tables also report the exact memory allocations of each model, allowing for a comparison between groups sizes.}
\label{tab:baseline_main}
\resizebox{\textwidth}{!}{%
\begin{tabular}{lrrrrrrrrrrrrrrrrrr}
\toprule
 &  &  & \multicolumn{8}{c}{C4 ↓ (Perplexity)} & \multicolumn{8}{c}{LM Eval Avg. ↑ (Accuracy over 8 zero-shot tasks)} \\
\cmidrule(lr){4-11}
\cmidrule(lr){12-19}
Method & Bits & Group & 1-7 & 1-13 & 1-30 & 2-7 & 2-13 & 2-70 & 3.1-8 & 3.1-70 & 1-7 & 1-13 & 1-30 & 2-7 & 2-13 & 2-70 & 3.1-8 & 3.1-70 \\
\midrule
Base & 16 & -- & 7.08 & 6.61 & 5.98 & 6.98 & 6.47 & 5.52 & 8.43 & 5.82 & 63.5 & 66.6 & 65.3 & 64.9 & 67.9 & 72.3 & 68.9 & 73.8 \\
\midrule
HQQ & 3 & 64 & 8.49 & 7.27 & 6.55 & 9.11 & 7.29 & 6.02 & 12.14 & 629.09 & 59.3 & 64.3 & 61.2 & 58.9 & 64.8 & 70.4 & 61.7 & 39.0 \\
HQQ & 3 & 128 & 9.36 & 7.60 & 6.75 & 10.17 & 7.69 & 6.48 & 14.46 & 620.89 & 58.4 & 64.1 & 61.5 & 56.8 & 63.4 & 69.5 & 58.2 & 34.1 \\
EntQuant & 3 & -- & \textbf{7.52} & \textbf{6.86} & \textbf{6.18} & \textbf{7.55} & \textbf{6.76} & \textbf{5.74} & \textbf{9.74} & \textbf{6.76} & \textbf{62.4} & \textbf{65.7} & \textbf{64.4} & \textbf{63.6} & \textbf{66.7} & \textbf{71.1} & \textbf{66.0} & \textbf{72.6} \\
\midrule
HQQ & 2 & 16 & 63.73 & 17.57 & 14.97 & 111.20 & 21.79 & 32.24 & 142.94 & 7.7e3 & 35.0 & 52.9 & 40.4 & 33.3 & 37.6 & 51.3 & 33.6 & 30.0 \\
HQQ & 2 & 32 & 1.4e3 & 186.58 & 60.96 & 1.2e3 & 214.22 & 323.93 & 1.9e3 & 3.0e4 & 30.6 & 32.9 & 34.0 & 29.9 & 30.3 & 31.9 & 29.9 & 29.9 \\
HQQ & 2 & 64 & 4.9e3 & 2.5e3 & 480.61 & 6.0e3 & 2.7e3 & 2.8e3 & 1.8e4 & 1.3e4 & 29.9 & 29.7 & 31.7 & 30.3 & 30.8 & 30.4 & 30.7 & 29.9 \\
EntQuant & 2.1 & -- & \textbf{9.90} & \textbf{8.16} & \textbf{7.25} & \textbf{10.59} & \textbf{7.95} & \textbf{6.47} & \textbf{17.56} & \textbf{9.92} & \textbf{37.5} & \textbf{56.3} & \textbf{60.4} & \textbf{57.5} & \textbf{63.1} & \textbf{67.9} & \textbf{52.6} & \textbf{68.6} \\
\midrule
EntQuant & 1.7 & -- & 17.76 & 11.19 & 9.40 & 27.92 & 11.14 & 8.43 & 135.24 & 7.9e3 & 41.1 & 34.1 & 54.0 & 43.7 & 50.3 & 60.2 & 36.8 & 35.5 \\
\bottomrule
\end{tabular}
}
\end{table*}

We first compare \methodname with two popular data-free compression methods: \amethod{HQQ}~\citep{badri2023hqq} and \amethod{NF4}~\citep{dettmers2023qlora}. \Cref{tab:baseline_main} presents results for the compression regime where \methodname excels, namely 2--3~bits. While \methodname already outperforms \amethod{HQQ} at 3-bit precision, the gap becomes substantial in the 2-bit regime, where all baseline methods exhibit functional collapse. The Pareto front plot in \Cref{fig:memory_vs_ppl} visualizes this observation across different compression levels.
Notably, at 2.1~bits, \methodname operates below the overhead introduced by a group size of 128, which yields 2.14~bits per parameter~\citep[Table~11]{chen-etal-2025-efficientqat}. To the best our knowledge, no other data-free method consistently performs adequately at effective 2-bit precision. The full version of \cref{tab:baseline_main} can be found in \cref{tab:baseline_full_llama1,tab:baseline_full_llama2,tab:baseline_full_llama31}, confirming that, at 4~bits, all methods perform similarly well with minor degradation. On the other hand, compressing significantly below 2~bits causes breakdowns, particularly for smaller base models.

\subsection{\methodnameplain Can Compete with Calibration and Fine-Tuning Methods}
\label{sec:experiments:literature}

\begin{table*}[t]
\centering
\caption{\methodnameplain vs.~calibration and fine-tuning methods on LLaMA-2 70B, comparing conceptual differences in subtable~(a) and accuracy in subtable~(b). Runtime specifications are taken from \citet{frantar2023gptq,shao2024omniquant,tseng2024quip,chen-etal-2025-efficientqat}, respectively. We report perplexity on C4 and WikiText-2, and avg.~accuracy over five zero-shot tasks; see \cref{tab:literature_full} for full results.
}
\label{tab:literature_main}
\begin{minipage}[t]{\columnwidth}
\strut\vspace*{-\baselineskip}\newline
{\centering\footnotesize
(a) \\[.25\baselineskip]
\scriptsize
\setlength{\tabcolsep}{3pt}
\begin{tabular*}{\linewidth}{l@{\extracolsep{\fill}}cccl}
\toprule
Method & Level & \shortstack{No\\Calibration} & \shortstack{No\\Training} & \shortstack{Compression Runtime\\Estimate} \\
\midrule
EntQuant      & 1 & \ding{51} & \ding{51}        & $<$10min (H100)$^\text{1}$ \\
GPTQ          & 2 & \ding{55} & \ding{51}        & 2--4h (A100-80GiB) \\
OmniQuant     & 2 & \ding{55} & \ding{51}        & 9--16h (A100-80GiB) \\
QuIP\#        & 3 & \ding{55} & \ding{55}$^\text{2}$ & $\sim$50h (8$\times$ A100-80GiB) \\
EfficientQAT  & 3 & \ding{51} & \ding{55}        & $\sim$41h (A100-80GiB) \\
\bottomrule
\end{tabular*}}
\scriptsize
\vspace{.25\baselineskip} \\
$^\text{1}$ This is a conservative estimate. Due to CPU buffering, the runtime of \methodnameplain also depends on the general hardware setup and utilization. Usually compression completes in significantly less than 10min and is even faster for smaller models. \\[.25\baselineskip]
$^\text{2}$ Recovery fine-tuning is optional but the default for QuIP\#.
\end{minipage}\hfill
\begin{minipage}[t]{\columnwidth}
\strut\vspace*{-\baselineskip}\newline
\centering\footnotesize
(b) \\[.25\baselineskip]
\scriptsize
\setlength{\tabcolsep}{3pt}
\begin{tabular*}{\linewidth}{l@{\extracolsep{\fill}}rrrrr}
\toprule
Method & Bits & Group & C4 ↓ & WikiText-2 ↓ & LM Eval Avg. ↑ \\
\midrule
Base & 16 & -- & 5.52 & 3.32 & 72.8 \\
\cmidrule{1-6}
EntQuant & 3 & -- & 5.74 & 3.62 & 71.7 (-1.6\%) \\
GPTQ & 3 & 128 & 5.85 & 3.85 & 71.5 (-1.9\%) \\
OmniQuant & 3 & 128 & 5.85 & 3.78 & 71.1 (-2.4\%) \\
QuIP\# & 3 & -- & 5.67 & 3.56 & 72.1 (-0.9\%) \\
EfficientQAT & 3 & 128 & 5.71 & 3.61 & 71.8 (-1.5\%) \\
\cmidrule{1-6}
EntQuant & 2.1 & -- & 6.47 & 4.52 & 68.6 (-5.8\%) \\
GPTQ & 2 & 128 & -- & -- & 34.4 (-52.8\%) \\
OmniQuant & 2 & 128 & 8.52 & 6.55 & 54.9 (-24.6\%) \\
QuIP\# & 2 & -- & 6.12 & 4.16 & 70.9 (-2.6\%) \\
EfficientQAT & 2 & 128 & 6.48 & 4.61 & 68.9 (-5.3\%) \\
\bottomrule
\end{tabular*}
\end{minipage}
\end{table*} 

We now compare \methodname to a variety of calibration- and fine-tuning-based quantization methods. These approaches constitute a fundamentally different category of algorithms (Level 2--4 in the taxonomy of \citealp{nagel2019datafree}), typically requiring (specialized) data access and substantial compute that may not be available to end users. \Cref{tab:literature_main} shows that \methodname performs on par with state-of-the-art fine-tuning-based methods like \amethod{QuIP\#} \citep{tseng2024quip} and \amethod{EfficientQAT} \citep{chen-etal-2025-efficientqat}, with only a slight gap at 2-bit precision. Full results are provided in \cref{tab:literature_full}, including more comparison methods.

We emphasize that the above study considers only base models and relatively basic benchmarks. For instruction-tuned or reasoning models and more complex tasks, it remains unclear how calibration or recovery fine-tuning can be performed without degrading specialized capabilities. 
Prior work has documented that even well-tuned calibration-based quantization methods often underperform on instruction-following and hallucination detection tasks \citep{10.24963/ijcai.2025/902}, and that low-bit quantization can degrade the capabilities of reasoning models \citep{liu_quantization_2025}.
This stands in stark contrast to \methodname, as we demonstrate in the next subsection.

\subsection{\methodnameplain Excels on Instruction-Tuned Models}
\label{sec:experiments:instruct}

As highlighted in \cref{fig:figure1}, we have evaluated \methodname in practical scenarios with models that end users would deploy. \methodname maintains strong performance on challenging benchmarks even at 2-bit precision, while 3-bit compression incurs negligible performance drops. Our evaluation covers instruction-following (IFEval), mathematical reasoning (GSM8K CoT), scientific reasoning (GPQA), and broad knowledge (MMLU).
\Cref{tab:instruct_full} presents full results across different base model sizes. 

Consistent with prior observations \cite{10.24963/ijcai.2025/902}, smaller models exhibit larger performance degradation under compression. These results underscore the plug-and-play nature of \methodname: requiring no data or model-specific adaptations, our approach avoids the catastrophic failures that plague other methods when applied to realistic scenarios.

\subsection{\methodnameplain Has Acceptable Inference Speed}
\label{sec:experiments:inference}

\begin{figure}[t]
    \centering
    \includegraphics[width=\linewidth]{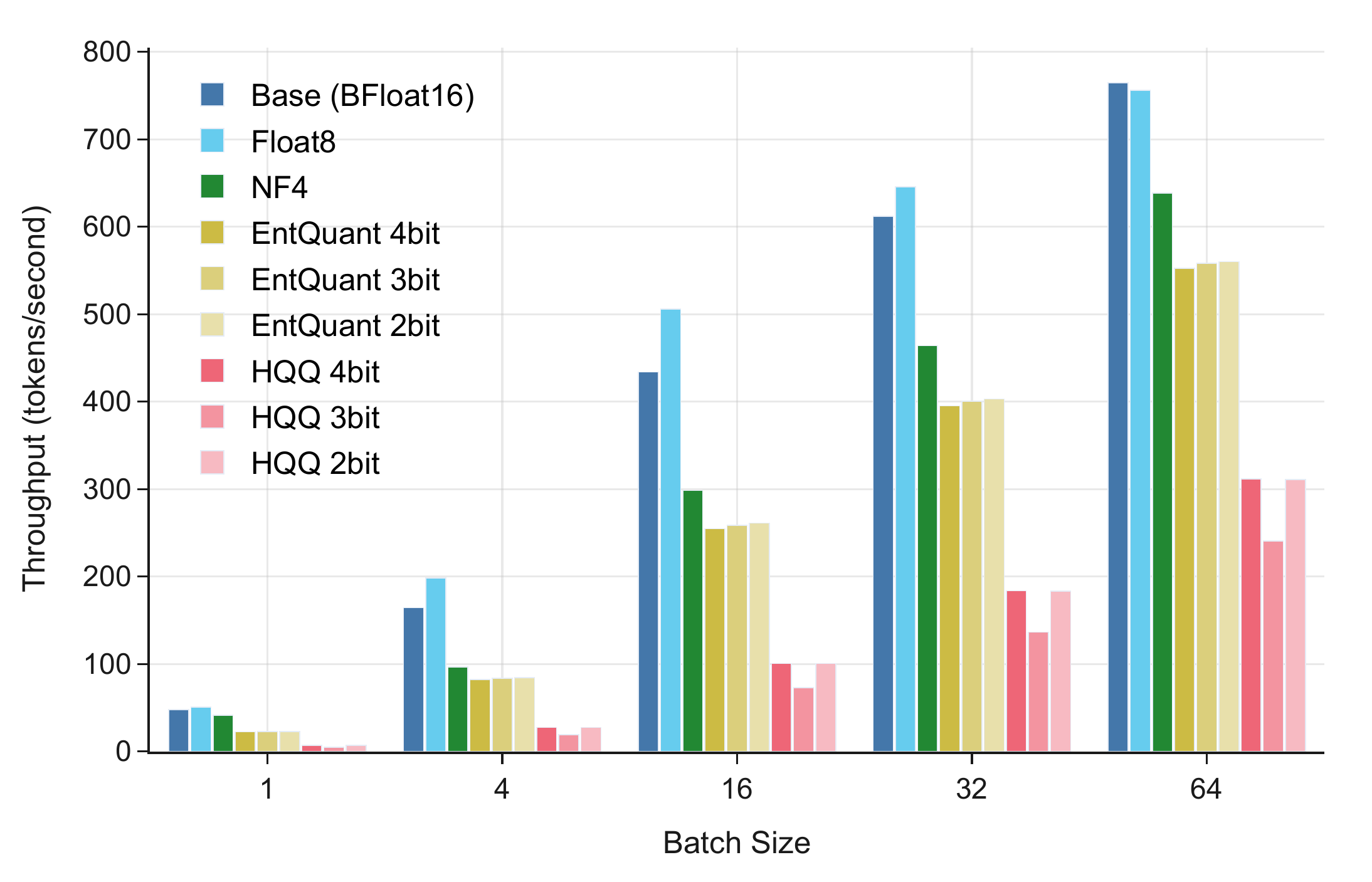}
    \caption{Inference throughput for LLaMA-2 13B in a standard prefill-decoding setting (input context length 512 tokens and 256 tokens generated). See \cref{fig:efficiency_full_throughput,fig:efficiency_full_latency,fig:efficiency_full_peak_memory} for more results.}
    \label{fig:efficiency_main}
    \vspace{-.5\baselineskip}
\end{figure}

\begin{figure}[ht]
    \centering
    \includegraphics[width=\linewidth]{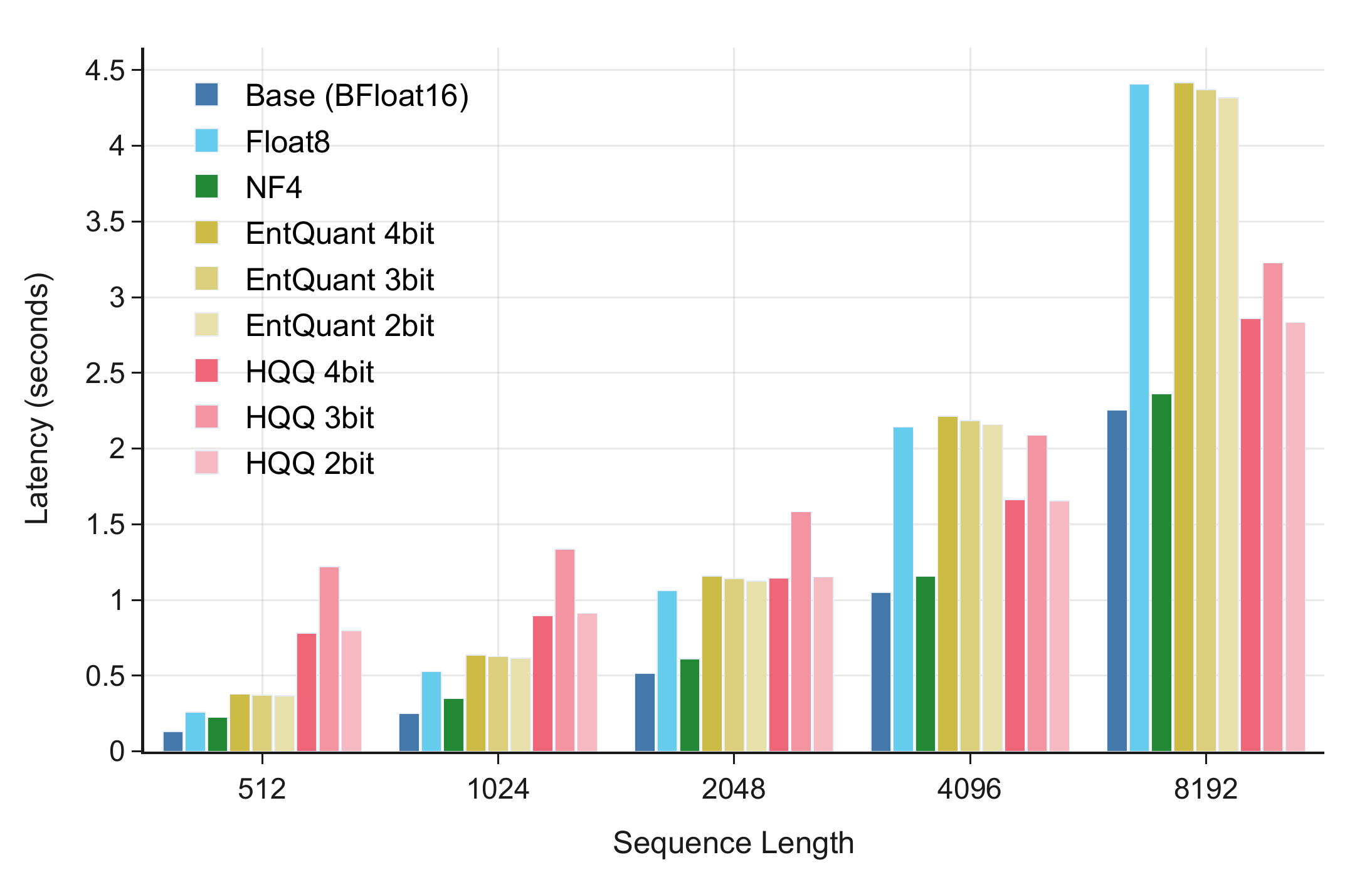}
    \caption{Inference latency for prefill on LLaMA-3.1 70B at batch size~1 across different input lengths. Since weights are decompressed once per transformer block per forward pass, the relative gap between \methodname and the \floateight \amethod{Marlin} kernel on which it builds shrinks as prefill length grows, illustrating that the decompression overhead is amortized over longer contexts. Note that for pure prefill, the \bfloat baseline is faster than \floateight and \methodname because the Marlin kernel's advantage only manifests in decoding regimes (see \cref{fig:efficiency_full_latency}).}
    \label{fig:efficiency_prefill_llama31_70b_latency}
    \vspace{-.5\baselineskip}
\end{figure}

A key difference between \methodname and standard quantization is the on-the-fly decompression of model weights (see \cref{fig:inference_pipeline,alg:decoding}). It is therefore crucial to verify whether this overhead remains acceptable. \Cref{fig:efficiency_main} shows that \methodname is only 1.5--2$\times$ slower than the \bfloat baseline, essentially matching \amethod{NF4} inference speed while \amethod{HQQ} lags behind. This overhead is in line with previous implementations of entropy coding of \bfloat weights \citep{zhang_70_2025, hao2024neuzipmemoryefficienttraininginference}.
Our GEMM implementation builds on the \floateight \amethod{Marlin} kernel \citep{frantar_marlin_2025}, denoted \floateight in \cref{fig:efficiency_main}, enabling a direct analysis of the decoding overhead. 
For reference, CPU offloading is approximately 3$\times$ slower for prefill and 45$\times$ slower for decoding than \methodname.\footnote{Tested on LLaMA-2 7B (prefill: batch size 8, seq.~len. 2048 / decoding: batch size 4, context len.~512, 64 tokens generated).}
Extended results for the full LLaMA-2 model family across different inference hyperparameter setups are provided in \cref{fig:efficiency_full_throughput,fig:efficiency_full_latency,fig:efficiency_full_peak_memory}. 
In particular, \cref{fig:efficiency_full_peak_memory} confirms that peak memory gains are most significant at batch size one, and a 70B parameter model can fit into a consumer GPU (32GiB RTX 5090) at 3~bits and lower, depending on the inference load.
In addition, decompression cost is independent of sequence length, since weights are decompressed once per transformer block per forward pass. Longer contexts therefore amortize the overhead. \Cref{fig:efficiency_prefill_llama31_70b_latency} confirms this on LLaMA-3.1 70B at batch size~1 across prefill lengths from 512 to 8192 tokens.

\subsection{Super Weights Help With Int8 Quantization}
\label{sec:experiments:sw}

\begin{figure}[t]
    \centering
    \includegraphics[width=\linewidth]{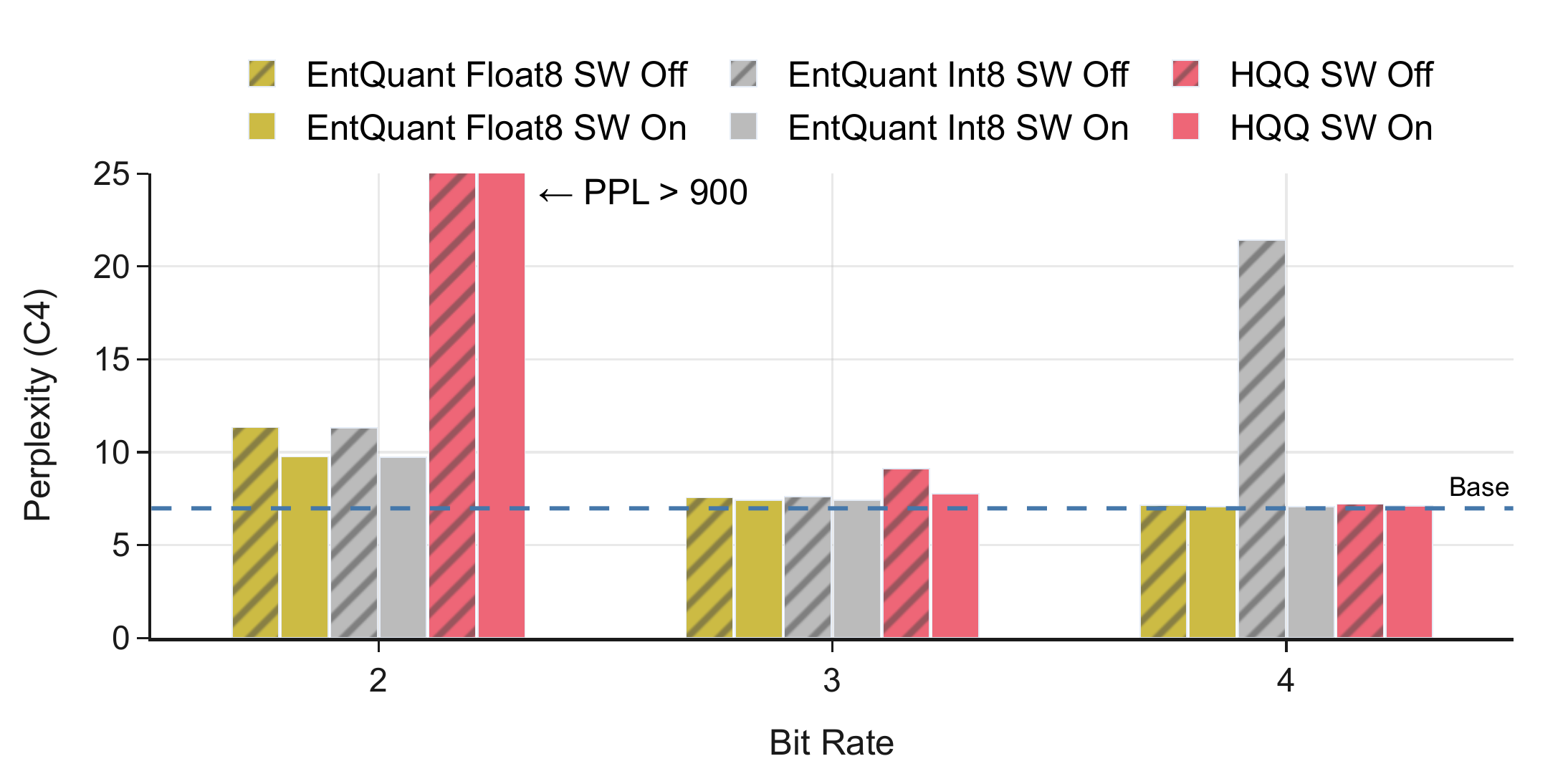}
    \caption{Comparison between \methodnameplain with \floateightplain base format (default) and \inteightplain for LLaMA-2 7B. Excluding layers with super weights (SW On) can substantially improve results for \inteightplain and modestly benefits \floateightplain. For HQQ, perplexity still explodes in either case at 2 bits. See \cref{tab:sw_ablation} for full results.}
    \label{fig:sw_ablation}
    \vspace{-.5\baselineskip}
\end{figure}

While the default version of \methodname uses \floateight, \inteight serves as a viable alternative base format. \Cref{fig:sw_ablation} shows that \inteight achieves similar compression results to \floateight but exhibits a certain sensitivity to so-called \emph{super weights}~\citep{yu2025superweightlargelanguage}. 
\citet{yu2025superweightlargelanguage} identified that typically fewer than 10 particularly large outliers, occurring predominantly in early down-projection layers, cause significant performance drops when erased. They also propose a simple and efficient detection algorithm that requires only a single (CPU) forward pass.

\Cref{fig:sw_ablation} demonstrates that precisely excluding layers containing super weights recovers expected performance for \inteight. Importantly, this improvement extends beyond \methodname: \cref{tab:sw_ablation} shows that accounting for super weights benefits \amethod{NF4} and \amethod{HQQ} as well. Note that excluding certain layers slightly affects the overall compression ratio; for \methodname, the reported entropy always accounts for this overhead by computing it over all linear layers. 
Moreover, we found that super weight handling also improves \floateight results for certain models, see \cref{app:implementation:sw} for implementation details and model-specific thresholds.

\subsection{Results for Float8 Activations (W8A16 vs. W8A8)}
\label{sec:experiments:a8}

Weight 8-bit quantization naturally invites consideration of activation quantization. Using dynamic quantization in \amethod{Quanto}, we evaluate W8A8 configurations. \Cref{tab:a8_ablation_pivoted} shows that dynamic quantization introduces a slight but acceptable performance drop, with LLaMA-2 70B exhibiting a somewhat larger gap than its smaller counterparts. Unfortunately, \amethod{Quanto} does not provide fused kernels for W8A8, precluding the evaluation of potential inference speedups.

\section{Related Work}
\label{sec:related}

\paragraph{Classical Roots in Signal Processing.}
The separation of discretization (quantization) from efficient representation (entropy coding) is a foundational principle in classical information theory and signal processing \citep{shannon_mathematical_1948, cover2006elements}. For decades, standards like \amethod{JPEG} \citep{wallace1991jpeg} have employed a pipeline where high-fidelity signals are first quantized and then losslessly compressed using Huffman or Arithmetic Coding. 
\methodname can be viewed as the modern realization of this classical pipeline for Large Foundation Models: rather than quantizing \amethod{DCT} coefficients as in \amethod{JPEG}, we quantize weight matrices, replacing static codebooks with high-throughput, parallel entropy (de)coders on the GPU. In addition, we directly optimize the rate-distortion function in the context of quantization, an approach also known as \emph{entropy-constrained quantization} \citep{chou_entropy-constrained_1989, Gray1998Quantization}.

\begin{table}[t]
\centering
\caption{Weight-only quantization (W8A16) versus combined weight and activation quantization (W8A8) for the LLaMA-2 base model family (7B, 13B, 70B). We report perplexity on C4.}
\label{tab:a8_ablation_pivoted}
\resizebox{\linewidth}{!}{%
\begin{tabular}{llrrrrrr}
\toprule
 &  & \multicolumn{2}{c}{2-7} & \multicolumn{2}{c}{2-13} & \multicolumn{2}{c}{2-70} \\
\cmidrule(lr){3-4}
\cmidrule(lr){5-6}
\cmidrule(lr){7-8}
Method & Bits & W8A16 & W8A8 & W8A16 & W8A8 & W8A16 & W8A8 \\
\midrule
Float8 & 8 & 6.99 & 7.25 & 6.48 & 6.65 & 5.53 & 6.46 \\
\midrule
EntQuant & 3.9 & 7.14 & 7.28 & 6.55 & 6.68 & 5.59 & 7.10 \\
EntQuant & 3 & 7.55 & 7.76 & 6.76 & 6.88 & 5.74 & 7.35 \\
EntQuant & 2 & 11.33 & 11.59 & 8.10 & 8.23 & 6.59 & 7.90 \\
\bottomrule
\end{tabular}
}
\vspace{-.5\baselineskip}
\end{table}

\paragraph{Entropy Coding in Neural Networks.}
In the deep learning era, \emph{Deep Compression} \citep{han2016deep} successfully instantiated entropy coding as a useful tool for model compression after quantization of small models e.g., AlexNet \citep{NIPS2012_c399862d} or VGG-16 \citep{simonyan_very_2015}. However, entropy coding was still an afterthought for additional storage benefits beyond quantization and pruning, without considering it for \emph{on-the-fly} decoding. With the recent advances in entropy coding on GPUs, \citet{zhang_70_2025} and \citet{hao2024neuzipmemoryefficienttraininginference} demonstrated that it can be integrated directly into the inference pipeline. Both approaches exploit the low entropy of exponent bits in \bfloat to achieve sizable storage reductions with manageable latency overhead. We go beyond these approaches by explicitly optimizing quantized weights to have much lower entropy.

\paragraph{Fixed Bit-width in Post-Training Quantization.}
Current LLM quantization methods have largely shifted to favoring fixed bit-widths, where the compression rate is strictly dictated by the underlying data type of quantized weights. Popular methods like \amethod{GPTQ} \citep{frantar2023gptq} and \amethod{AWQ} \citep{lin2024awq} are highly effective at 4-bit precision with \intfour but seem to approach a fundamental barrier at lower bit-rates \citep{egiazarian2024aqlm}. This limitation arises from the rigid coupling of storage costs to the bit-width, which restricts the expressiveness of these methods. \methodname breaks this limitation by turning to the classical entropy-constrained paradigm: using a high-precision data type (e.g., \floateight) to preserve signal quality while achieving low-bit storage costs (e.g., 2.1~bits per parameter) via entropy coding (see \cref{tab:unique_val}).

\paragraph{Extreme Quantization ($<$ 3~bits).}
To push beyond 3~bits, recent state-of-the-art PTQ methods like \amethod{AQLM} \citep{egiazarian2024aqlm} and \amethod{QuIP\#} \citep{tseng2024quip} employ complex vector quantization combined with incoherence processing (e.g., randomized Hadamard transforms). 
Alternatively, quantization-aware training methods like \amethod{EfficientQAT} \citep{chen-etal-2025-efficientqat} achieve competitive results at 2--3~bits through end-to-end training of quantization parameters.
While effective, these Level 3--4 approaches introduce significant complexity and are not data-free, making transfer to specialized instruction-tuned or reasoning models non-trivial.
Moreover, achieving competitive extreme compression results typically requires the explicit handling of outliers \citep{dettmers2024spqr, kim2024squeezellm},
which is not case for \methodname.

\section{Discussion}
\label{sec:conclusion}

Standard quantization couples compression rate to weight precision, so current methods either fail at extreme compression levels or require costly recovery training. \methodname breaks this paradigm by decoupling weight precision from compression rate through entropy coding, achieving arbitrary sizes down to effective 2~bits per parameter while retaining robust model performance without calibration data or retraining. To our knowledge, \methodname is the first method to achieve functional extreme compression in a purely data-free manner, suggesting that entropy coding can overcome the performance barriers that fixed bit-width quantization is approaching.

\paragraph{Limitations.}
We intentionally designed \methodname to be the simplest version of the ``Entropy Coding Meets Quantization'' framework. In our pursuit of accessibility, we opted for data-free compression with simplified $\ell_1$-based regularization, foregoing complex grouping structures or custom fused operations. Many extensions are conceivable, including more sophisticated quantization schemes, advanced entropy proxies, and fused decoding kernels. While our implementation uses NVIDIA's \amethod{nvCOMP} library, \amethod{ANS} is a commoditized technology available across platforms (e.g., \amethod{hipANS} for AMD), ensuring broad hardware compatibility.

Our evaluation study was conducted with a budget of 5K GPU hours, in which we selected a reasonable set of benchmarks and model sizes. A more extensive evaluation on real-world tasks and larger, mixture-of-experts models is planned for future work. Since \methodname operates on weight matrices alone, it is architecture-agnostic and applies beyond LLMs, e.g., to diffusion models~\citep{zhang_70_2025}.

Although our implementation matches the inference speed of \amethod{NF4} (\cref{fig:efficiency_main}), \methodname is currently slower than the uncompressed \bfloat baseline. The history of low-bit quantization is instructive here: at batch size~1, early implementations of GPTQ and QuIP were similarly slower than uncompressed inference (${\sim}0.6\times$ and ${\sim}0.4\times$ TPOT, respectively), only to overtake it by a wide margin (${\sim}2.9\times$ and ${\sim}3.2\times$) once dedicated fused kernels such as \amethod{Marlin}~\citep{frantar_marlin_2025} and \amethod{E8P}~\citep{tseng2024quip} were introduced. \methodname's overhead sits squarely in this pre-kernel-optimization range, and fused decompression-GEMM kernels are a natural avenue for closing the gap. Moreover, its decoding is highly parallel and compute-intensive, aligning with the continuing hardware trend where GPU compute capabilities outpace memory bandwidth growth.

\paragraph{Final Remarks.}
In self-hosting scenarios, memory, not inference speed, is typically the binding constraint, and users are often willing to trade latency for model quality. \methodname directly addresses this sweet spot: by accepting a modest slowdown, it enables deployment of much larger models that would otherwise exceed memory limits. Moreover, compressing a 70B model takes less than 10 minutes without data or domain knowledge, enabling immediate adoption of newly released models.

\clearpage
\section*{Acknowledgements}

The authors thank Felix Möller for helpful discussions and feedback. 
We thank Jannis Klinkenberg and Fritz Niesel for their technical cluster support.

We kindly acknowledge funding by the European Union -- NextGenerationEU -- and the German Federal Ministry for Economic Affairs and Energy within the project ``Souveräne KI für Europa (SOOFI)'' (grant no. 13IPC040H).
Computational resources were provided by the German AI Service Center WestAI and used to conduct the numerical experiments of this work.

\section*{Impact Statement}

By substantially reducing the memory required to deploy large language models, \methodname lowers hardware costs and broadens access to powerful models for researchers and practitioners with limited computational resources.
On the other hand, lowering the barrier to deployment may also facilitate the use of large models in settings where adequate safety guardrails are not in place.
Future work should evaluate the interaction between extreme compression and safety-relevant model behaviors, such as toxicity and harmful-content generation.

\bibliography{references_manual}
\bibliographystyle{icml2026}

\newpage
\clearpage
\appendix
\onecolumn

\counterwithin{figure}{section}
\counterwithin{table}{section}
\counterwithin{equation}{section}

\section{Implementation Details}
\label{app:implementation}

\subsection{Implementation of \methodname}
\label{app:implementation:entquant}

The quantization and optimization components of \methodname described in \cref{sec:method:entquant} use the \amethod{Optimum Quanto} package as backend. 
During inference with \floateight weights (specifically \texttt{torch.float8e4m3}), we employ the \amethod{Marlin} kernel \citep{frantar_marlin_2025} in the forward pass of quantized linear layers, which is natively supported by \amethod{Quanto}.

\begin{figure}[ht]
    \centering
    \includegraphics[width=.55\linewidth]{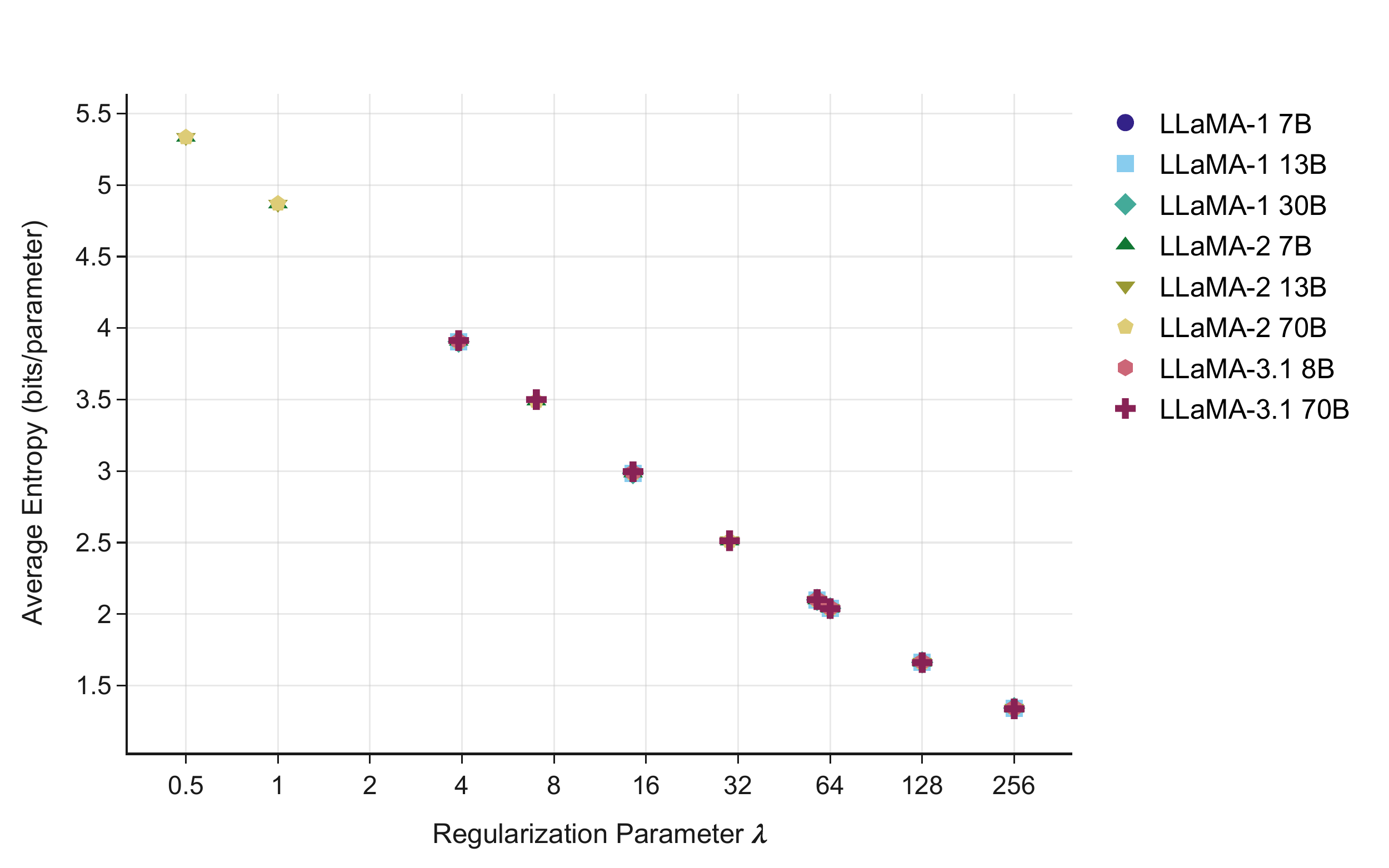}
    \caption{Regularization parameter $\lambda$ in \eqref{eq:optimization} vs.~resulting average entropy for all \methodnameplain models considered in \cref{tab:baseline_main}. The almost perfect clustering shows that the choices of $\lambda$ are model-independent, which drastically simplifies the hyperparameter selection process.}
    \label{fig:entropy_vs_reg_param}
\end{figure}

The regularization parameter $\lambda$ in \eqref{eq:optimization} maps to a target entropy rate. While this mapping is non-linear, it is strictly monotone and robust across all linear layers and models considered; see \cref{fig:entropy_vs_reg_param} for empirical evidence. Consequently, the set of $\lambda$ hyperparameters was globally selected to match our grid of target entropies. For the learning rate of \amethod{L-BFGS}, we use $0.25$ when $\lambda > 30$ and $1.0$ when $\lambda \leq 30$, but in general, we found that \amethod{L-BFGS} is quite robust to the choice of its hyperparamters.

For \floateight, we resolve signed zeros to eliminate unnecessary redundancy in the representation.

The block-wise decompression pipeline is described in \cref{sec:method:entquant}.
The \amethod{ANS} (de)compression component of \methodname uses NVIDIA's \amethod{nvCOMP} package (version 5.1.0) with a chunk-size of 256KiB \citep{nvcomp}.
While each weight matrix is optimized individually via \eqref{eq:optimization}, Line~4 of \cref{alg:encoding} simply concatenates all flattened attention and projection matrices into a large vector, as detailed in the main text.

\begin{figure}[ht]
    \centering
    \includegraphics[width=.85\linewidth]{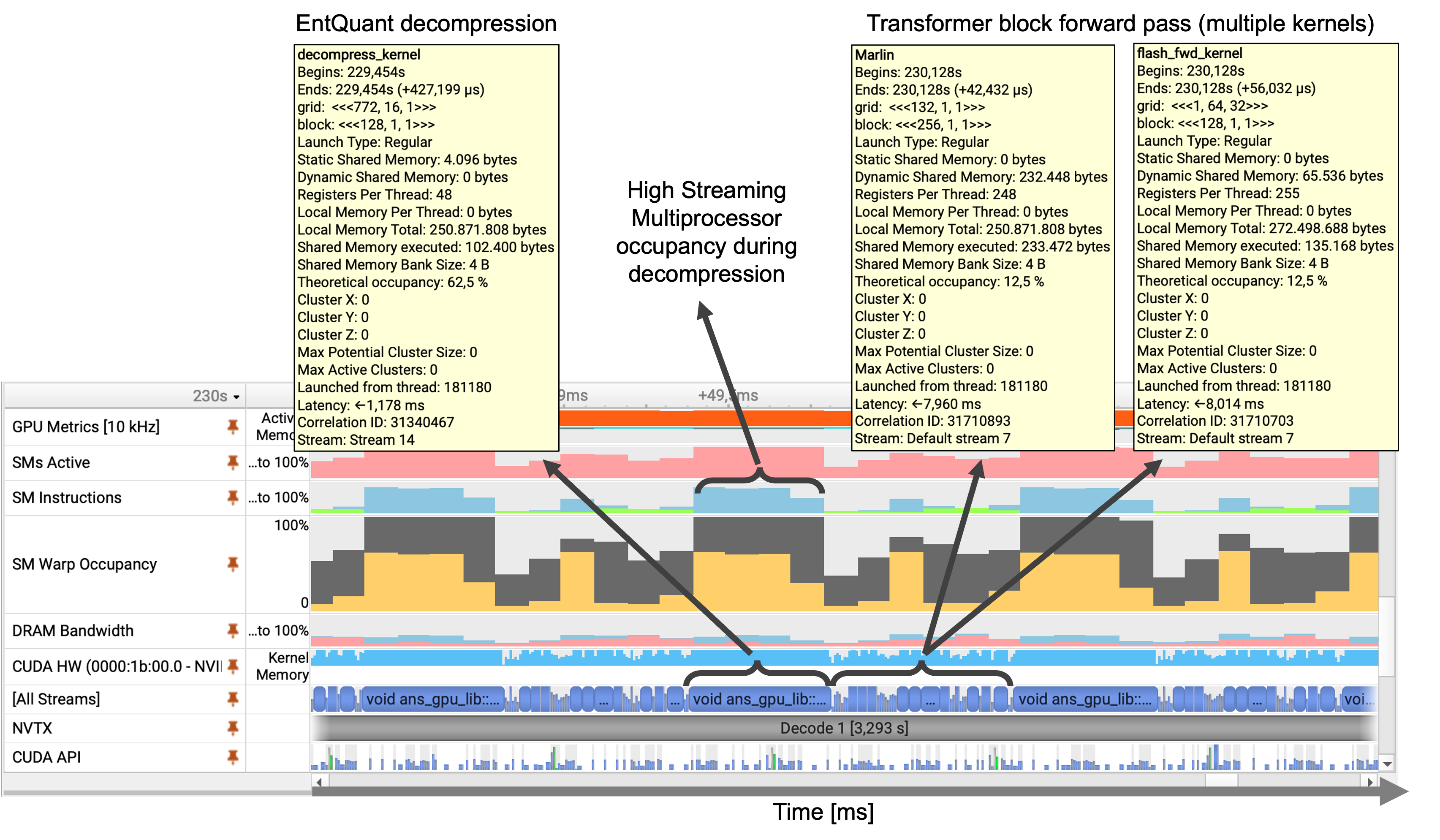}
    \caption{NVIDIA Nsight™ Systems profiling of \methodnameplain inference (decoding) on LLaMA-2 7B showing the interleaving of ANS decompression and forward pass of each transformer block. The latter is a series of multiple forward passes of linear layers, involving kernels like Marlin \citep{frantar_marlin_2025} and FlashAttention \citep{dao2022flashattention,dao2023flashattention2}.}
    \label{fig:nsight}
\end{figure}

GPU profiling (\cref{fig:nsight}) confirms full utilization during decompression with dense alternation between decompression and forward kernels. This analysis also indicates optimization potential for multi-GPU settings. For example, upcoming blocks could be decompressed by another (idle) GPUs during the current forward pass. We leave such further optimization steps to future work.

Finally, we note that \amethod{nvCOMP}'s implementation of \amethod{ANS} operates on a byte-level. Using \floateight or \inteight as a base format is therefore a natural choice and leads to near-optimal entropy-based encoding.

\newpage

\subsection{Super Weights}
\label{app:implementation:sw}

Adopting the algorithm of~\citet{yu2025superweightlargelanguage}, we detect super weights via a single CPU forward pass with a dummy prompt, considering only down-projection layers as candidates. Activation thresholds were manually selected per model family:
\begin{itemize}
    \item LLaMA-1 7B--30B: threshold $= 50$
    \item LLaMA-3.1 8B Base \& Instruct: threshold $= 50$  
    \item Qwen3 8B--32B: threshold $= 200$
    \item All other models: threshold $= \infty$ (no super weight exclusions)
\end{itemize}
Excluded down-projection layers are still quantized to 8~bits and compressed with \amethod{ANS}, yielding an entropy of approximately 6.5~bits per parameter.

\subsection{Baseline Methods}
\label{app:implementation:baseline}

All experiments use \amethod{PyTorch} \citep{Ansel_PyTorch_2_Faster_2024} and HuggingFace's \amethod{Transformers} library \citep{wolf-etal-2020-transformers}. All base models are loaded and evaluated in \bfloat precision.  Analogously to \methodname, we use \amethod{Optimum Quanto} for \floateight quantization, employing the \amethod{Marlin} kernel \citep{frantar_marlin_2025}, see also~\cref{fig:nsight}. For \amethod{NF4} \citep{dettmers2023qlora} and \amethod{HQQ} \citep{badri2023hqq}, we rely on the official \amethod{Transformers} integrations, which are based on the \amethod{BitsAndBytes} and \amethod{HQQ} packages, respectively.
The results for the comparison methods reported in \cref{tab:literature_main} and \cref{tab:literature_full} are taken from the literature and were not generated in-house, see the captions for more details.

\subsection{Evaluations tasks}
\label{app:implementation:eval}

\paragraph{Perplexity.} We follow standard implementations for computing perplexity on C4 and WikiText-2. Context length is 2048 for LLaMA-1/2 and Qwen3, and 4096 for all other models.

\paragraph{LM-Eval Benchmarks.} We use the EleutherAI LM Evaluation Harness~\citep{eval-harness} (package version 0.4.9.2) with thinking mode disabled and chat templates enabled for instruct models. Default generation settings and unmodified chat templates/system prompts are used throughout. \Cref{tab:eval_tasks} below summarizes our settings for all considered evaluation tasks.

\begin{table}[ht]
\centering
\caption{Evaluation tasks and configurations used in this work.}
\label{tab:eval_tasks}
\small
\begin{tabular}{lllll}
\toprule
Benchmark & Reference & LM-Eval Identifier & Metric & Shots \\
\midrule
ARC-Easy & \citep{allenai:arc} & \texttt{arc\_easy} & acc & 0 \\
ARC-Challenge & \citep{allenai:arc} & \texttt{arc\_challenge} & acc & 0 \\
HellaSwag & \citep{zellers2019hellaswag} & \texttt{hellaswag} & acc & 0 \\
WinoGrande & \citep{winogrande} & \texttt{winogrande} & acc & 0 \\
PIQA & \citep{Bisk2020} & \texttt{piqa} & acc & 0 \\
BoolQ & \citep{clark-etal-2019-boolq} & \texttt{boolq} & acc & 0 \\
OpenbookQA & \citep{mihaylov-etal-2018-suit} & \texttt{openbookqa} & acc & 0 \\
LAMBADA & \citep{paperno-etal-2016-lambada} & \texttt{lambada\_openai} & acc & 0 \\
GSM8K CoT & \citep{cobbe2021gsm8k} & \texttt{gsm8k\_cot} & exact\_match, flexible-extract & 8 \\
GPQA Main & \citep{rein2024gpqa} & \texttt{gpqa\_main\_n\_shot} & acc & 5 \\
MMLU & \citep{hendryckstest2021} & \texttt{mmlu} & acc & 5 \\
IFEval & \citep{zhou2023instructionfollowing} & \texttt{ifeval} & prompt\_level\_strict\_acc & 0 \\
\bottomrule
\end{tabular}
\end{table}

\subsection{Testbed Hardware}
\label{app:implementation:hardware}

All evaluations were performed on a SLURM-based HPC cluster with NVIDIA H100 GPUs. All considered \methodname models fit on a single H100 during evaluation. \bfloat base models exceeding single-GPU capacity were distributed across 2--3 GPUs using the dispatch framework of HuggingFace's \amethod{accelerate} package.

\clearpage

\section{More Conceptual Insights on \methodname}
\label{app:insights}

\subsection{Low Entropy Is Linked to Sparsity}
\label{app:sparsity}

\begin{figure}[ht]
    \centering
    \includegraphics[width=.55\linewidth]{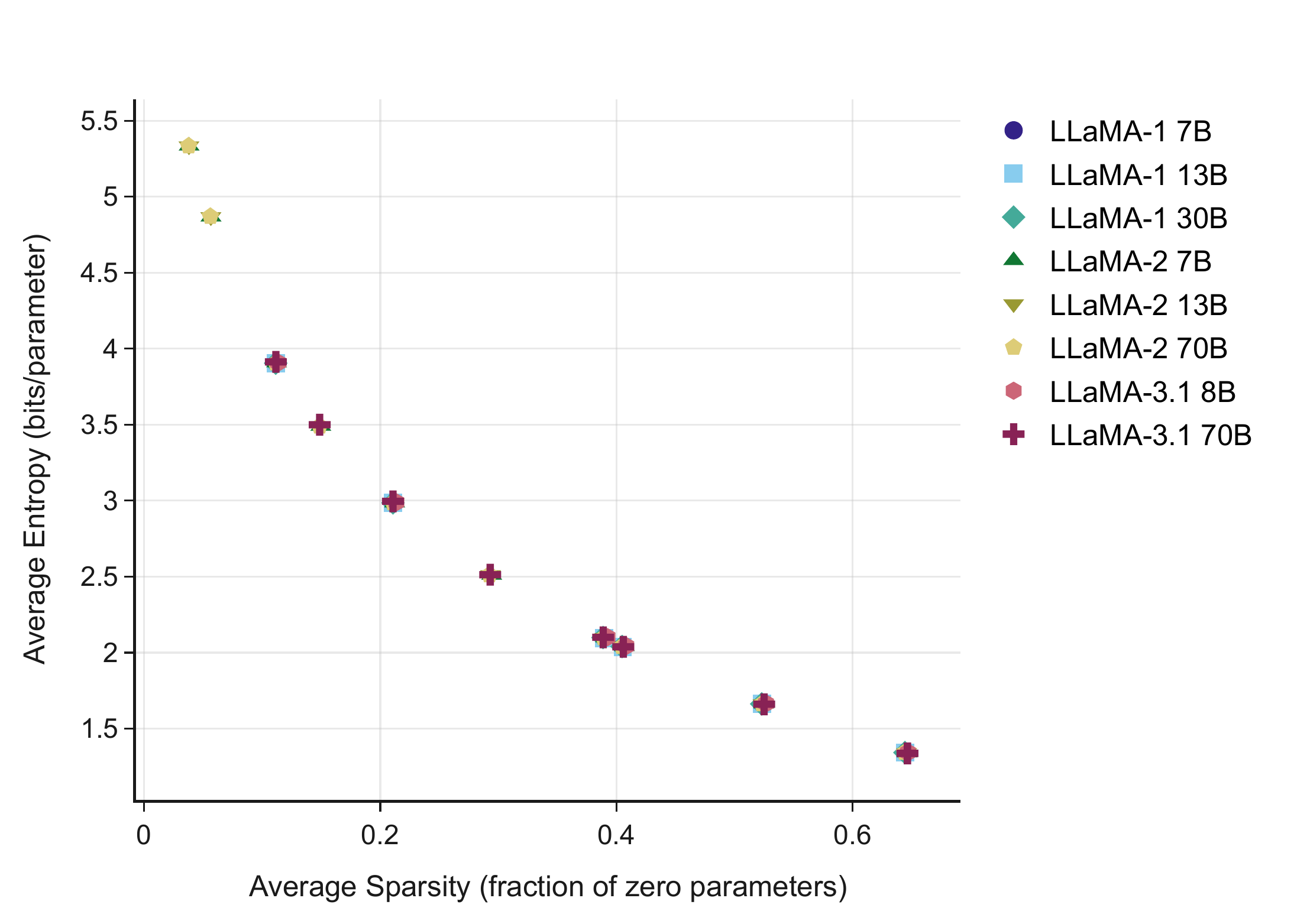}
    \caption{Total sparsity over all linear layers vs.~average entropy for all \methodnameplain models considered in \cref{tab:baseline_main}. The almost perfect clustering indicates a model-independent relationship between sparsity and entropy. Hence, to a certain extent, \methodnameplain also operates as an (unstructured) ``soft pruning'' method whose compression step can also leverage non-zero weights.}
    \label{fig:entropy_vs_sparsity}
\end{figure}

\paragraph{Sparsity alone does not explain \methodname's entropy reductions.}
To quantify the role of sparsity, we consider baselines where a fraction $s$ of weights are zeroed. The resulting weight distribution is a mixture of a point mass at zero (probability~$s$) and a non-zero component with conditional entropy $H_{\mathrm{nz}}$, giving total entropy
\begin{equation*}
H(s) = -s\log_2 s - (1-s)\log_2(1-s) + (1-s) \cdot H_{\mathrm{nz}},
\end{equation*}
where the first two terms account for the zero/non-zero indicator. We instantiate $H_{\mathrm{nz}}$ in two ways: (1)~uniform over 255 \floateightplain levels, $H_{\mathrm{nz}} = \log_2 255 \approx 7.99$; (2)~the original ${\sim}6.5$-bit non-zero distribution (\cref{fig:entropy_vs_sparsity}).
\Cref{tab:sparsity_entropy} shows that both baselines require sparsity $> 0.8$ to reach 2~bits, whereas \methodname achieves 2~bits at sparsity $\approx 0.4$.
This demonstrates that, under symmetric weight quantization, $\ell_1$ regularization reshapes the non-zero weight distribution toward lower entropy beyond the pure sparsity effect.
\Cref{tab:unique_val} further confirms this: at 2 effective bits, \methodname retains ${\sim}35$ unique values versus only 4 for fixed 2-bit quantization, preserving the expressivity that prevents functional collapse (\cref{tab:baseline_main}).

\begin{table}[h]
\centering
\caption{Theoretical entropy (bits per parameter) as a function of sparsity $s$ under two baselines.}
\label{tab:sparsity_entropy}
\small
\begin{tabular}{lccccccccc}
\toprule
& $s{=}0.1$ & $s{=}0.2$ & $s{=}0.3$ & $s{=}0.4$ & $s{=}0.5$ & $s{=}0.6$ & $s{=}0.7$ & $s{=}0.8$ & $s{=}0.9$ \\
\midrule
Uniform non-zeros & 7.66 & 7.12 & 6.48 & 5.77 & 5.00 & 4.17 & 3.28 & 2.32 & 1.27 \\
Original distribution & 6.32 & 5.92 & 5.43 & 4.87 & 4.25 & 3.57 & 2.83 & 2.02 & 1.12 \\
\bottomrule
\end{tabular}
\end{table}

\begin{figure}[h]
    \centering
    \begin{subfigure}[t]{0.48\linewidth}
        \centering
        \includegraphics[width=\linewidth]{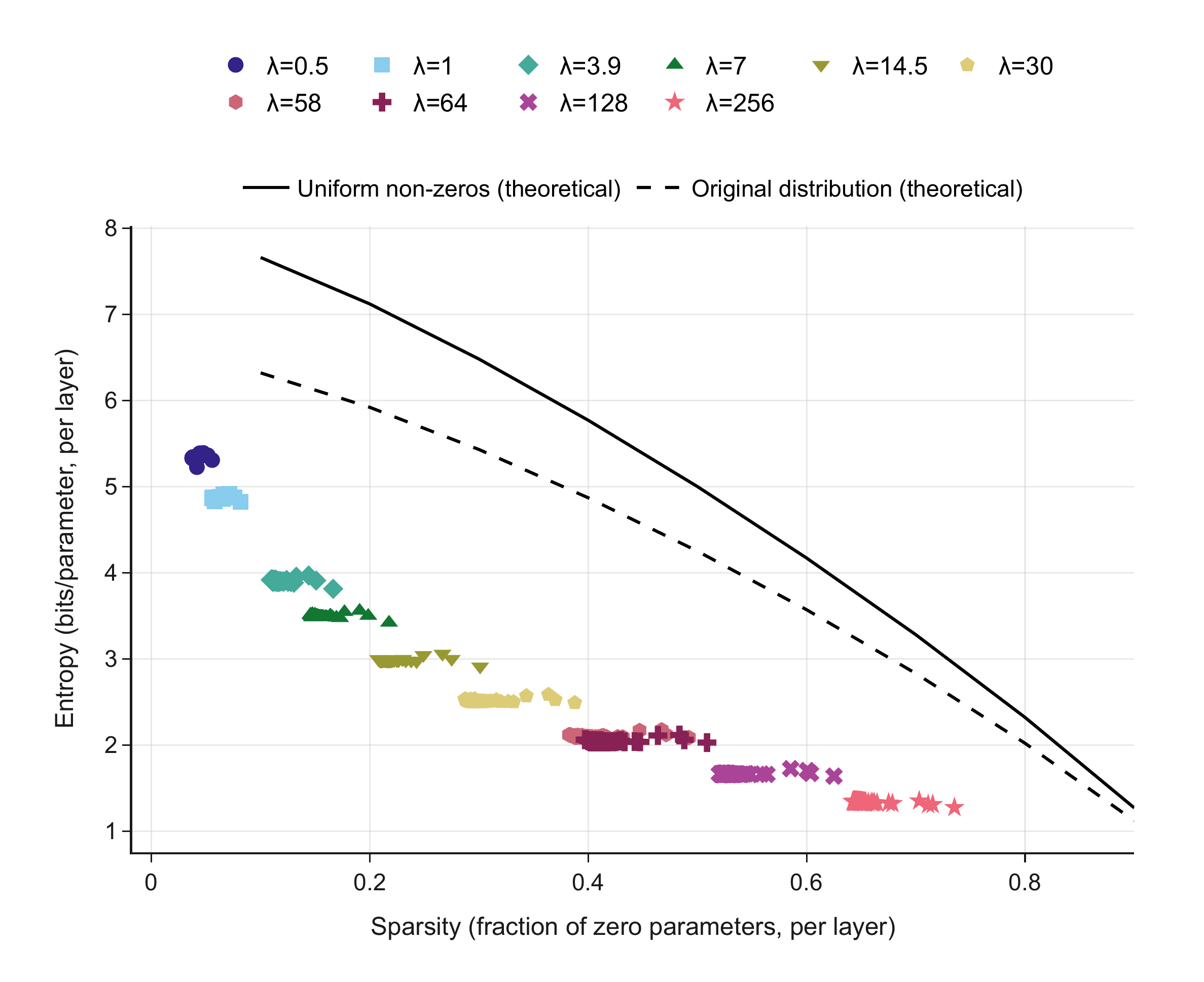}
        \caption{Layer entropy vs.\ sparsity.}
    \end{subfigure}
    \hfill
    \begin{subfigure}[t]{0.48\linewidth}
        \centering
        \includegraphics[width=\linewidth]{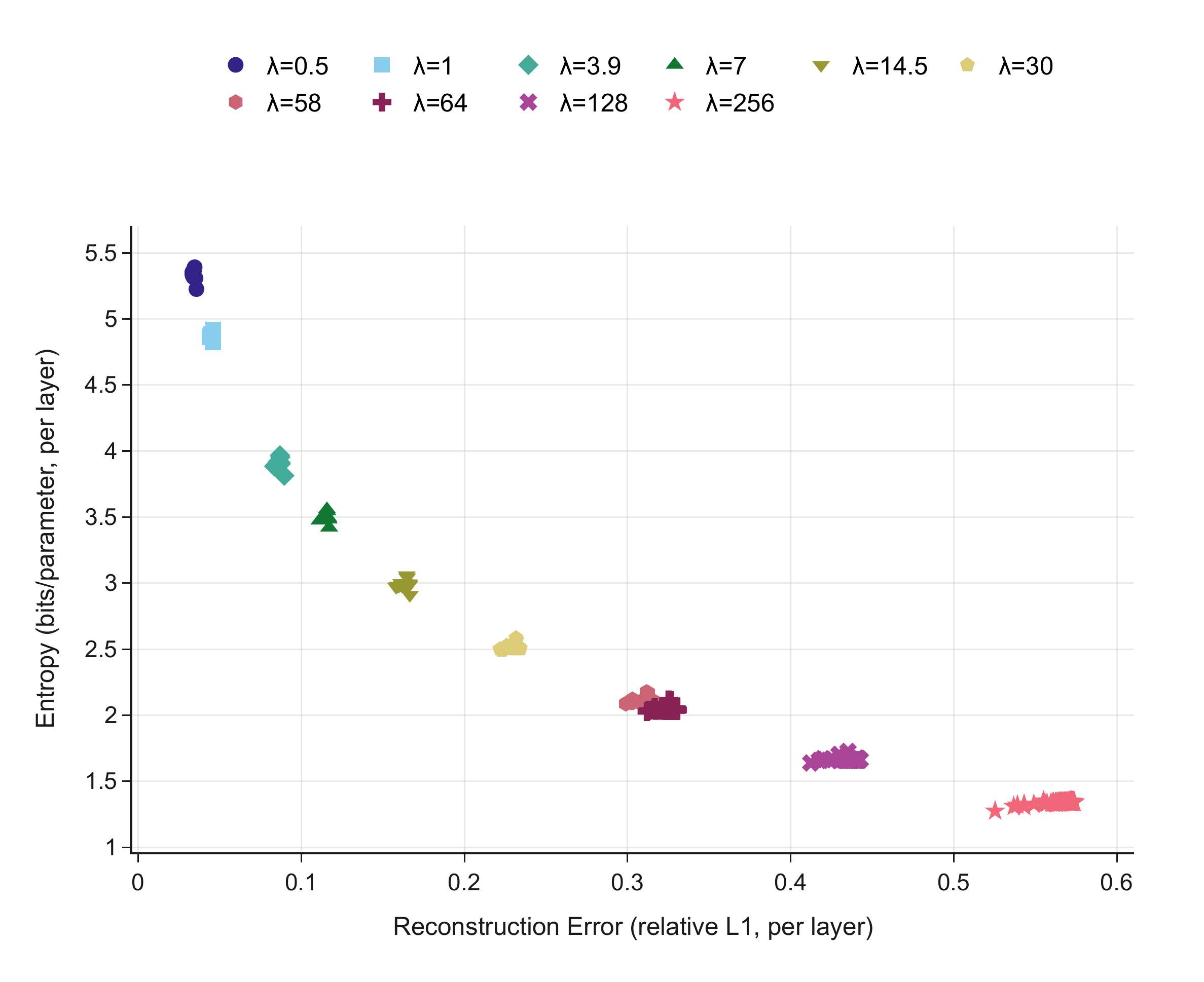}
        \caption{Layer entropy vs.\ reconstruction error (relative $\ell_1$).}
    \end{subfigure}
    \caption{Layer-wise scatter plots for \floateight-quantized LLaMA-2 7B across a sweep of the entropy-regularization strength $\lambda$. Each marker is one transformer weight matrix, where color/shape encodes $\lambda$. \emph{Left:} per-layer entropy against per-layer sparsity, with the theoretical entropy bounds for a uniform non-zero distribution (solid) and the original \floateight value distribution (dashed) overlaid. \emph{Right:} per-layer entropy against per-layer relative $\ell_1$ reconstruction error. Both relations are tight and monotone across the layer population, confirming that the global $\lambda$ knob induces a consistent rate--distortion trade-off at the level of individual layers.}
    \label{fig:per_layer_entropy}
\end{figure}

\paragraph{Per-layer entropy and reconstruction error.}
\Cref{fig:per_layer_entropy} shows layer-wise scatter plots for LLaMA-2 7B, relating per-layer entropy to sparsity and to weight matrix reconstruction error.
We caution against correlating single-layer entropy with downstream accuracy, as layers interact non-linearly during inference, making it difficult to attribute accuracy changes to an individual layer's entropy in isolation.

\subsection{Entropy Control via a Surrogate Regularization Term}
\label{app:entropy_regularization}

In this section, we derive a basic argument that demonstrates how the minimization of a regularization term based on the $\ell_1$-norm of a random matrix allows for control of its entropy.
In accordance with the framework presented in \cref{sec:method}, we focus on matrix entries with discrete distributions, which we assume to be defined over $\mathbb{Z}$.
Let $X$ be a scalar random variable on $\mathbb{Z}$. The \emph{maximum entropy principle} \citep[Example 5.2]{Polyanskiy_Wu_2025} shows that for all $\lambda > 0$, we have
\begin{equation}
    \label{eq:max-entropy}
    H(X) \leq \lambda \, \mathbb{E}[|X|] + \log Z(\lambda),
\end{equation}
where
\begin{equation*}
    Z(\lambda) := \sum_{n \in \mathbb{Z}} \exp(-\lambda |n|) = 2\sum_{n=0}^\infty \exp(-\lambda n) - 1 = \frac{2}{1 - \exp(-\lambda)} -1.
\end{equation*}
We note that this result is sharp, i.e., the bound can not be improved and is attained for a known class of Gibbs distributions.
We apply this result to a random matrix $\mathbf{W}_q = (\mathbf{W}_q^{(i,j)})$ with $1 \leq i \leq M$, $1 \leq j \leq N$.
Using entropy subaddivity, we have
\begin{align*}
    H(\mathbf{W}_q) & \leq \sum_{ij} H(\mathbf{W}_q^{(i,j)}) \\
    &\leq \sum_{ij} \left(\lambda \, \mathbb{E}\left[ \left|\mathbf{W}_q^{(i,j)}\right|\right] + \log Z(\lambda) \right) \\
    & =\lambda \, \mathbb{E}\left[\left\Vert \mathbf{W}_q \right\Vert_{1}\right] + MN  \log Z(\lambda) \quad \text{for all } \lambda \geq 0,
\end{align*}
where we used \eqref{eq:max-entropy} in the second step.
The nonlinear relationship between $H(\mathbf{W}_q)$ and $\left\Vert \mathbf{W}_q \right\Vert_{1}$ is captured in the tradeoff in $\lambda$ in the above sum, as we have $\lim_{\lambda \to \infty} MN  \log Z(\lambda) = 0$.
The basic intuition in the regularization context is as follows: If we assume that for some $\delta > 0$, a minimization procedure for the regularizer $\lambda R(\mathbf{W}_q) = \lambda ||\mathbf{W}_q||_1$ over $\mathbf{W}_q$ has led to the result 
\begin{equation*}
    \lambda \mathbb{E}\left[
        R(\mathbf{W}_q)
    \right]
    \leq \delta,
\end{equation*}
we have a guaranteed corresponding entropy control in terms of
\begin{equation*}
    H(\mathbf{W}_q) 
    \leq \delta +
    MN  \log Z(\lambda).
\end{equation*}
This derivation also shows that the tightest worst-case control of the entropy obtained through minimization of $R(\mathbf{W}_q)$ for fixed $\lambda$ and $\delta$ depends on the ambient dimension $MN$.

\clearpage

\section{Full Results for Data-Free Methods}
\label{app:results_baseline}

\begin{table*}[h!]
\centering
\caption{Full baseline results for the LLaMA-1 base model family (7B, 13B, 30B). We report perplexity on C4 and WikiText-2, as well as zero-shot accuracy on eight tasks from LM Eval. All results are generated in-house using the configurations described in \cref{app:implementation}. Best results per model and bit-rate group are in \textbf{bold}. This table extends the results presented in \cref{tab:baseline_main}.}
\label{tab:baseline_full_llama1}
\resizebox{\textwidth}{!}{%
\begin{tabular}{llrrrrrrrrrrrrrr}
\toprule
Model & Method & Bits & Group & Size (GiB) & C4 ↓ & WikiText-2 ↓ & ARC-E ↑ & ARC-C ↑ & HellaSwag ↑ & WinoGrande ↑ & PIQA ↑ & BoolQ ↑ & OBQA ↑ & LAMBADA ↑ & LM Eval Avg. ↑ \\
\midrule
LLaMA-1 7B & Base & 16 & -- & 12.6 & 7.08 & 5.68 & 74.5 & 41.6 & 56.4 & 69.4 & 78.1 & 73.6 & 40.6 & 73.9 & 63.5 \\
\cmidrule{2-16}
 & NF4 & 4 & 64 & 3.9 & 7.25 & 5.82 & 74.2 & 40.9 & \textbf{55.8} & 68.2 & 77.4 & 73.1 & \textbf{42.4} & \textbf{73.4} & \textbf{63.2} \\
 & HQQ & 4 & 64 & 3.9 & 7.32 & 5.91 & 72.7 & 39.4 & 53.0 & 66.4 & \textbf{77.7} & 73.3 & 38.6 & 70.7 & 61.5 \\
 & HQQ & 4 & 128 & \textbf{3.7} & 7.37 & 5.96 & 73.8 & 40.6 & 55.3 & \textbf{68.5} & 77.4 & \textbf{73.6} & 41.6 & 72.4 & 62.9 \\
 & EntQuant & 3.9 & -- & \textbf{3.7} & \textbf{7.18} & \textbf{5.78} & \textbf{74.4} & \textbf{41.0} & 55.4 & 68.4 & 77.6 & 73.0 & 39.8 & 72.5 & 62.8 \\
\cmidrule{2-16}
 & EntQuant & 3.5 & -- & 3.4 & 7.27 & 5.88 & 74.5 & 41.1 & 56.6 & 69.1 & 78.6 & 75.0 & 42.6 & 72.8 & 63.8 \\
\cmidrule{2-16}
 & HQQ & 3 & 64 & 3.3 & 8.49 & 6.95 & 70.2 & 36.9 & 52.5 & 66.1 & 75.8 & 69.0 & 37.2 & 66.9 & 59.3 \\
 & HQQ & 3 & 128 & \textbf{3.1} & 9.36 & 7.67 & 69.8 & 36.8 & 49.3 & 64.6 & 74.5 & \textbf{71.8} & 35.8 & 64.3 & 58.4 \\
 & EntQuant & 3 & -- & \textbf{3.1} & \textbf{7.52} & \textbf{6.11} & \textbf{71.8} & \textbf{38.9} & \textbf{55.2} & \textbf{69.4} & \textbf{77.5} & \textbf{71.8} & \textbf{41.8} & \textbf{72.4} & \textbf{62.4} \\
\cmidrule{2-16}
 & EntQuant & 2.6 & -- & 2.7 & 8.10 & 6.59 & 71.7 & 37.0 & 52.8 & 64.8 & 74.7 & 72.7 & 37.8 & 68.2 & 60.0 \\
\cmidrule{2-16}
 & HQQ & 2 & 16 & 3.5 & 63.73 & 68.43 & 28.9 & 21.2 & 25.8 & 49.1 & 58.5 & \textbf{61.4} & 26.4 & 8.4 & 35.0 \\
 & HQQ & 2 & 32 & 2.8 & 1.4e3 & 3.2e3 & 25.3 & 20.6 & 25.4 & 48.7 & 52.7 & 45.8 & 26.4 & 0.0 & 30.6 \\
 & HQQ & 2 & 64 & 2.4 & 4.9e3 & 7.7e3 & 25.0 & 21.8 & \textbf{25.9} & \textbf{49.6} & 53.1 & 38.9 & 24.8 & 0.0 & 29.9 \\
 & EntQuant & 2.1 & -- & \textbf{2.3} & \textbf{9.90} & \textbf{8.34} & \textbf{30.1} & \textbf{22.8} & \textbf{25.9} & 48.3 & \textbf{64.8} & 50.5 & \textbf{26.6} & \textbf{31.0} & \textbf{37.5} \\
\cmidrule{2-16}
 & EntQuant & 1.7 & -- & 2.1 & 17.76 & 15.67 & 49.7 & 25.8 & 34.9 & 52.9 & 65.0 & 49.9 & 27.2 & 23.6 & 41.1 \\
 & EntQuant & 1.4 & -- & 1.8 & 396.08 & 700.35 & 28.4 & 20.8 & 27.4 & 50.4 & 55.2 & 45.7 & 25.2 & 0.1 & 31.7 \\
\midrule
LLaMA-1 13B & Base & 16 & -- & 24.2 & 6.61 & 5.09 & 77.1 & 46.2 & 59.8 & 72.8 & 79.1 & 77.8 & 44.4 & 75.9 & 66.6 \\
\cmidrule{2-16}
 & NF4 & 4 & 64 & 7.3 & 6.71 & 5.20 & 76.4 & 45.8 & 59.1 & 71.1 & 78.5 & \textbf{79.1} & \textbf{46.2} & \textbf{75.8} & \textbf{66.5} \\
 & HQQ & 4 & 64 & 7.3 & 6.72 & 5.21 & 76.3 & 46.0 & 59.2 & 71.8 & 78.7 & 77.5 & 44.0 & 75.1 & 66.1 \\
 & HQQ & 4 & 128 & 6.9 & 6.75 & 5.25 & 75.9 & \textbf{46.2} & 59.0 & 72.0 & 78.7 & 77.6 & 45.0 & 75.0 & 66.2 \\
 & EntQuant & 3.9 & -- & \textbf{6.8} & \textbf{6.67} & \textbf{5.15} & \textbf{76.6} & 45.4 & \textbf{59.5} & \textbf{72.2} & \textbf{79.1} & 76.7 & 44.8 & 75.6 & 66.2 \\
\cmidrule{2-16}
 & EntQuant & 3.5 & -- & 6.2 & 6.72 & 5.21 & 76.2 & 45.1 & 59.2 & 72.2 & 78.8 & 77.0 & 44.8 & 75.4 & 66.1 \\
\cmidrule{2-16}
 & HQQ & 3 & 64 & 6.1 & 7.27 & 5.71 & 75.8 & 43.3 & 56.9 & 69.6 & 77.3 & 75.2 & 42.6 & 73.4 & 64.3 \\
 & HQQ & 3 & 128 & 5.7 & 7.60 & 6.03 & 75.5 & 43.0 & 55.8 & 69.8 & 77.8 & 75.7 & 42.6 & 72.9 & 64.1 \\
 & EntQuant & 3 & -- & \textbf{5.5} & \textbf{6.86} & \textbf{5.36} & \textbf{76.5} & \textbf{44.7} & \textbf{59.0} & \textbf{71.5} & \textbf{78.5} & \textbf{76.9} & \textbf{44.0} & \textbf{74.2} & \textbf{65.7} \\
\cmidrule{2-16}
 & EntQuant & 2.6 & -- & 4.8 & 7.19 & 5.63 & 76.8 & 42.5 & 57.6 & 70.2 & 78.0 & 75.4 & 44.8 & 72.7 & 64.7 \\
\cmidrule{2-16}
 & HQQ & 2 & 16 & 6.5 & 17.57 & 13.10 & \textbf{65.9} & 31.4 & 44.9 & \textbf{61.8} & 72.1 & 65.3 & \textbf{36.2} & 45.7 & 52.9 \\
 & HQQ & 2 & 32 & 5.0 & 186.58 & 198.94 & 35.7 & 20.6 & 27.9 & 50.3 & 57.1 & 38.6 & 25.6 & 7.1 & 32.9 \\
 & HQQ & 2 & 64 & 4.3 & 2.5e3 & 4.7e3 & 26.2 & 21.1 & 26.1 & 49.1 & 53.5 & 37.9 & 23.6 & 0.0 & 29.7 \\
 & EntQuant & 2.1 & -- & \textbf{4.1} & \textbf{8.16} & \textbf{6.46} & 65.7 & \textbf{35.6} & \textbf{51.8} & 60.0 & \textbf{73.6} & \textbf{70.6} & 34.2 & \textbf{59.0} & \textbf{56.3} \\
\cmidrule{2-16}
 & EntQuant & 1.7 & -- & 3.6 & 11.19 & 10.04 & 28.5 & 20.3 & 26.4 & 51.2 & 59.2 & 54.5 & 23.4 & 9.3 & 34.1 \\
 & EntQuant & 1.4 & -- & 3.1 & 69.31 & 87.73 & 35.5 & 22.3 & 32.6 & 51.1 & 57.1 & 50.7 & 24.6 & 5.4 & 34.9 \\
\midrule
LLaMA-1 30B & Base & 16 & -- & 60.6 & 5.98 & 4.10 & 75.9 & 45.1 & 61.3 & 66.1 & 77.9 & 81.0 & 40.2 & 74.7 & 65.3 \\
\cmidrule{2-16}
 & NF4 & 4 & 64 & 17.6 & 6.08 & 4.22 & 75.5 & 44.5 & 60.8 & 67.2 & 77.8 & 81.5 & 38.8 & 74.3 & 65.1 \\
 & HQQ & 4 & 64 & 17.6 & 6.08 & 4.21 & 75.5 & 45.6 & \textbf{61.3} & 67.4 & \textbf{78.3} & \textbf{82.2} & \textbf{40.6} & \textbf{74.5} & \textbf{65.7} \\
 & HQQ & 4 & 128 & 16.7 & 6.09 & 4.23 & 75.6 & 44.3 & 61.2 & 66.8 & 78.0 & 81.4 & 38.8 & 73.8 & 65.0 \\
 & EntQuant & 3.9 & -- & \textbf{16.2} & \textbf{6.03} & \textbf{4.18} & \textbf{76.2} & \textbf{46.8} & 61.1 & \textbf{67.5} & 77.9 & 81.7 & 38.6 & 74.2 & 65.5 \\
\cmidrule{2-16}
 & EntQuant & 3.5 & -- & 14.7 & 6.07 & 4.22 & 75.5 & 44.0 & 61.0 & 66.9 & 78.2 & 80.3 & 40.0 & 73.7 & 65.0 \\
\cmidrule{2-16}
 & HQQ & 3 & 64 & 14.6 & 6.55 & 4.82 & 71.2 & 38.5 & 56.9 & \textbf{66.2} & 76.7 & 74.3 & \textbf{37.8} & 67.8 & 61.2 \\
 & HQQ & 3 & 128 & 13.7 & 6.75 & 5.06 & 72.5 & 41.3 & 57.2 & 63.5 & 75.6 & 77.4 & 35.2 & 69.3 & 61.5 \\
 & EntQuant & 3 & -- & \textbf{12.8} & \textbf{6.18} & \textbf{4.41} & \textbf{75.3} & \textbf{45.1} & \textbf{60.5} & 65.6 & \textbf{77.3} & \textbf{81.0} & \textbf{37.8} & \textbf{72.8} & \textbf{64.4} \\
\cmidrule{2-16}
 & EntQuant & 2.6 & -- & 11.1 & 6.46 & 4.80 & 74.1 & 42.2 & 59.5 & 63.8 & 77.7 & 79.4 & 36.6 & 70.3 & 63.0 \\
\cmidrule{2-16}
 & HQQ & 2 & 16 & 15.7 & 14.97 & 13.96 & 43.8 & 19.7 & 36.4 & 51.2 & 63.7 & 59.4 & 29.2 & 19.6 & 40.4 \\
 & HQQ & 2 & 32 & 12.0 & 60.96 & 83.88 & 32.7 & 20.6 & 28.5 & 49.6 & 56.7 & 50.4 & 27.2 & 6.5 & 34.0 \\
 & HQQ & 2 & 64 & 10.1 & 480.61 & 790.32 & 27.6 & 21.8 & 26.1 & 48.5 & 54.0 & 48.9 & 26.0 & 0.5 & 31.7 \\
 & EntQuant & 2.1 & -- & \textbf{9.3} & \textbf{7.25} & \textbf{5.52} & \textbf{70.1} & \textbf{36.4} & \textbf{56.3} & \textbf{65.0} & \textbf{74.8} & \textbf{75.4} & \textbf{38.0} & \textbf{67.3} & \textbf{60.4} \\
\cmidrule{2-16}
 & EntQuant & 1.7 & -- & 7.9 & 9.40 & 7.71 & 60.7 & 30.2 & 51.5 & 61.2 & 70.8 & 68.3 & 33.2 & 56.3 & 54.0 \\
 & EntQuant & 1.4 & -- & 6.7 & 78.66 & 36.24 & 40.2 & 24.5 & 35.8 & 52.2 & 59.9 & 57.7 & 27.8 & 13.1 & 38.9 \\
\bottomrule
\end{tabular}
}
\end{table*}
\begin{table*}[t]
\centering
\caption{Full baseline results for the LLaMA-2 base model family (7B, 13B, 70B). We report perplexity on C4 and WikiText-2, as well as zero-shot accuracy on eight tasks from LM Eval. The 6.5-bit rate reported for \floateightplain results from a lossless compression step as in \methodnameplain. All results are generated in-house using the configurations described in \cref{app:implementation}. Best results per model and bit-rate group are in \textbf{bold}. This table extends the results presented in \cref{tab:baseline_main}.}
\label{tab:baseline_full_llama2}
\resizebox{\textwidth}{!}{%
\begin{tabular}{llrrrrrrrrrrrrrr}
\toprule
Model & Method & Bits & Group & Size (GiB) & C4 ↓ & WikiText-2 ↓ & ARC-E ↑ & ARC-C ↑ & HellaSwag ↑ & WinoGrande ↑ & PIQA ↑ & BoolQ ↑ & OBQA ↑ & LAMBADA ↑ & LM Eval Avg. ↑ \\
\midrule
LLaMA-2 7B & Base & 16 & -- & 12.6 & 6.98 & 5.47 & 75.5 & 42.5 & 57.1 & 69.7 & 77.7 & 79.2 & 44.4 & 73.4 & 64.9 \\
\cmidrule{2-16}
 & Float8 & 6.5 & -- & 5.7 & 6.99 & 5.48 & 75.8 & 43.1 & 57.2 & 69.0 & 78.1 & 79.1 & 44.4 & 73.5 & 65.0 \\
\cmidrule{2-16}
 & EntQuant & 5.3 & -- & 4.8 & 7.00 & 5.50 & 75.5 & 42.9 & 57.2 & 69.9 & 77.7 & 79.3 & 44.6 & 72.8 & 65.0 \\
\cmidrule{2-16}
 & EntQuant & 4.9 & -- & 4.4 & 7.04 & 5.54 & 76.2 & 42.2 & 57.1 & 69.8 & 78.0 & 79.1 & 43.8 & 73.4 & 64.9 \\
\cmidrule{2-16}
 & NF4 & 4 & 64 & 3.9 & 7.16 & 5.65 & 75.5 & \textbf{43.2} & 56.8 & 69.9 & 77.7 & 78.8 & 44.4 & 72.5 & \textbf{64.9} \\
 & HQQ & 4 & 64 & 3.9 & 7.20 & 5.68 & \textbf{75.8} & 42.1 & 56.8 & \textbf{70.1} & \textbf{78.0} & \textbf{80.0} & 44.0 & 72.3 & \textbf{64.9} \\
 & HQQ & 4 & 128 & \textbf{3.7} & 7.24 & 5.74 & 75.6 & 42.9 & 56.7 & 68.8 & \textbf{78.0} & 77.1 & 43.4 & \textbf{72.9} & 64.4 \\
 & EntQuant & 3.9 & -- & \textbf{3.7} & \textbf{7.14} & \textbf{5.62} & 75.0 & 42.7 & \textbf{57.2} & 68.3 & \textbf{78.0} & 78.9 & \textbf{45.2} & 72.5 & 64.7 \\
\cmidrule{2-16}
 & EntQuant & 3.5 & -- & 3.4 & 7.26 & 5.73 & 76.1 & 42.5 & 56.4 & 69.1 & 77.5 & 78.4 & 45.2 & 72.6 & 64.7 \\
\cmidrule{2-16}
 & HQQ & 3 & 64 & 3.3 & 9.11 & 7.05 & 68.5 & 37.3 & 51.0 & 65.6 & 75.1 & 68.8 & 39.6 & 65.8 & 58.9 \\
 & HQQ & 3 & 128 & 3.1 & 10.17 & 7.99 & 67.2 & 33.8 & 47.4 & 65.5 & 73.3 & 67.1 & 38.0 & 61.8 & 56.8 \\
 & EntQuant & 3 & -- & \textbf{3.0} & \textbf{7.55} & \textbf{5.95} & \textbf{74.6} & \textbf{42.9} & \textbf{56.3} & \textbf{68.4} & \textbf{77.7} & \textbf{75.1} & \textbf{42.0} & \textbf{71.4} & \textbf{63.6} \\
\cmidrule{2-16}
 & EntQuant & 2.5 & -- & 2.7 & 8.35 & 6.57 & 74.5 & 40.6 & 55.0 & 68.0 & 76.6 & 71.6 & 41.6 & 68.1 & 62.0 \\
\cmidrule{2-16}
 & HQQ & 2 & 16 & 3.5 & 111.20 & 115.73 & 31.0 & 19.6 & 27.3 & 49.9 & 55.5 & 52.9 & 26.0 & 3.9 & 33.3 \\
 & HQQ & 2 & 32 & 2.8 & 1.2e3 & 1.6e3 & 26.8 & 21.7 & 26.0 & 49.5 & 54.1 & 37.8 & 23.0 & 0.0 & 29.9 \\
 & HQQ & 2 & 64 & 2.4 & 6.0e3 & 7.5e3 & 25.3 & 21.0 & 26.1 & 50.4 & 52.8 & 41.7 & 25.2 & 0.0 & 30.3 \\
 & EntQuant & 2.1 & -- & \textbf{2.3} & \textbf{10.59} & \textbf{8.29} & \textbf{68.9} & \textbf{36.2} & \textbf{51.6} & \textbf{65.7} & \textbf{73.7} & \textbf{67.9} & \textbf{42.2} & \textbf{54.0} & \textbf{57.5} \\
\cmidrule{2-16}
 & EntQuant & 1.7 & -- & 2.0 & 27.92 & 22.48 & 52.7 & 26.7 & 43.3 & 57.4 & 67.1 & 42.2 & 32.2 & 28.0 & 43.7 \\
 & EntQuant & 1.3 & -- & 1.8 & 2.4e3 & 2.6e3 & 27.4 & 21.2 & 26.7 & 49.3 & 54.4 & 40.9 & 26.4 & 0.2 & 30.8 \\
\midrule
LLaMA-2 13B & Base & 16 & -- & 24.2 & 6.47 & 4.88 & 78.9 & 47.2 & 60.2 & 72.2 & 79.7 & 82.6 & 45.6 & 76.5 & 67.9 \\
\cmidrule{2-16}
 & Float8 & 6.5 & -- & 10.8 & 6.48 & 4.89 & 78.9 & 47.4 & 60.3 & 72.3 & 79.3 & 82.0 & 45.2 & 76.3 & 67.7 \\
\cmidrule{2-16}
 & EntQuant & 5.3 & -- & 8.9 & 6.49 & 4.90 & 78.9 & 47.4 & 60.1 & 72.5 & 79.3 & 82.2 & 46.0 & 76.4 & 67.8 \\
\cmidrule{2-16}
 & EntQuant & 4.9 & -- & 8.2 & 6.50 & 4.91 & 79.0 & 46.9 & 60.1 & 72.4 & 78.9 & 82.4 & 45.6 & 76.5 & 67.7 \\
\cmidrule{2-16}
 & NF4 & 4 & 64 & 7.3 & 6.57 & 4.98 & \textbf{79.3} & 46.7 & 59.8 & \textbf{72.4} & \textbf{78.7} & \textbf{82.6} & 45.0 & 76.2 & \textbf{67.6} \\
 & HQQ & 4 & 64 & 7.3 & 6.57 & 4.98 & 79.0 & 46.7 & \textbf{59.9} & 72.3 & 78.3 & 81.3 & 44.6 & 75.8 & 67.2 \\
 & HQQ & 4 & 128 & 6.9 & 6.60 & 5.00 & 78.2 & 45.2 & 59.6 & 71.8 & 78.6 & 81.8 & \textbf{46.0} & \textbf{76.8} & 67.2 \\
 & EntQuant & 3.9 & -- & \textbf{6.8} & \textbf{6.55} & \textbf{4.97} & 78.1 & \textbf{46.8} & \textbf{59.9} & 72.3 & \textbf{78.7} & 82.5 & 45.0 & 76.5 & 67.5 \\
\cmidrule{2-16}
 & EntQuant & 3.5 & -- & 6.2 & 6.61 & 5.01 & 78.3 & 46.3 & 59.8 & 72.5 & 78.8 & 81.8 & 46.4 & 75.8 & 67.5 \\
\cmidrule{2-16}
 & HQQ & 3 & 64 & 6.1 & 7.29 & 5.60 & 76.0 & 41.9 & 57.3 & 69.8 & 77.2 & 80.5 & 42.8 & 72.9 & 64.8 \\
 & HQQ & 3 & 128 & 5.7 & 7.69 & 5.87 & 74.3 & 41.5 & 56.4 & 67.3 & 76.6 & 78.0 & 41.2 & 72.1 & 63.4 \\
 & EntQuant & 3 & -- & \textbf{5.4} & \textbf{6.76} & \textbf{5.15} & \textbf{77.3} & \textbf{44.3} & \textbf{59.4} & \textbf{72.1} & \textbf{78.5} & \textbf{82.0} & \textbf{45.2} & \textbf{74.6} & \textbf{66.7} \\
\cmidrule{2-16}
 & EntQuant & 2.5 & -- & 4.7 & 7.10 & 5.44 & 77.0 & 43.9 & 57.7 & 70.7 & 78.3 & 80.2 & 43.6 & 73.3 & 65.6 \\
\cmidrule{2-16}
 & HQQ & 2 & 16 & 6.5 & 21.79 & 17.73 & 42.4 & 21.8 & 33.3 & 50.0 & 60.1 & 56.5 & 24.6 & 12.4 & 37.6 \\
 & HQQ & 2 & 32 & 5.0 & 214.22 & 302.92 & 27.3 & 20.1 & 26.8 & 49.7 & 54.7 & 37.8 & 25.0 & 1.2 & 30.3 \\
 & HQQ & 2 & 64 & 4.3 & 2.7e3 & 3.8e3 & 26.6 & 19.8 & 26.0 & 49.2 & 52.2 & 46.8 & 25.6 & 0.0 & 30.8 \\
 & EntQuant & 2.1 & -- & \textbf{4.1} & \textbf{7.95} & \textbf{6.24} & \textbf{74.5} & \textbf{41.4} & \textbf{55.6} & \textbf{69.6} & \textbf{76.9} & \textbf{75.2} & \textbf{41.4} & \textbf{69.8} & \textbf{63.1} \\
\cmidrule{2-16}
 & EntQuant & 1.7 & -- & 3.5 & 11.14 & 9.25 & 60.8 & 30.0 & 46.6 & 57.7 & 69.7 & 57.2 & 31.0 & 49.3 & 50.3 \\
 & EntQuant & 1.3 & -- & 3.0 & 43.77 & 54.51 & 36.2 & 23.0 & 32.5 & 49.8 & 59.1 & 57.5 & 24.0 & 6.3 & 36.1 \\
\midrule
LLaMA-2 70B & Base & 16 & -- & 128.5 & 5.52 & 3.32 & 82.6 & 54.4 & 65.3 & 80.2 & 81.6 & 85.4 & 49.4 & 79.3 & 72.3 \\
\cmidrule{2-16}
 & Float8 & 6.5 & -- & 54.9 & 5.53 & 3.33 & 82.6 & 53.9 & 65.4 & 80.2 & 81.6 & 85.0 & 49.2 & 79.2 & 72.1 \\
\cmidrule{2-16}
 & EntQuant & 5.3 & -- & 44.6 & 5.53 & 3.33 & 82.6 & 54.0 & 65.3 & 80.3 & 81.6 & 85.1 & 49.2 & 79.4 & 72.2 \\
\cmidrule{2-16}
 & EntQuant & 4.9 & -- & 40.9 & 5.55 & 3.35 & 83.0 & 54.1 & 65.1 & 80.3 & 81.3 & 85.2 & 48.8 & 79.5 & 72.2 \\
\cmidrule{2-16}
 & NF4 & 4 & 64 & 36.8 & \textbf{5.59} & 3.42 & \textbf{82.5} & 53.6 & \textbf{64.9} & 79.1 & 81.3 & \textbf{85.0} & 48.2 & \textbf{79.0} & \textbf{71.7} \\
 & HQQ & 4 & 64 & 36.8 & 5.60 & 3.43 & 82.2 & 53.0 & 64.5 & \textbf{79.7} & 81.2 & 84.3 & \textbf{49.0} & 78.8 & 71.6 \\
 & HQQ & 4 & 128 & 34.8 & 5.63 & 3.46 & 81.9 & 53.3 & 64.5 & 79.3 & \textbf{81.4} & 84.7 & 48.8 & 78.7 & 71.6 \\
 & EntQuant & 3.9 & -- & \textbf{33.2} & \textbf{5.59} & \textbf{3.40} & 82.1 & \textbf{53.8} & 64.8 & 78.7 & 81.0 & \textbf{85.0} & 48.2 & 78.6 & 71.5 \\
\cmidrule{2-16}
 & EntQuant & 3.5 & -- & 29.9 & 5.65 & 3.48 & 82.0 & 52.8 & 64.4 & 78.5 & 81.0 & 83.7 & 48.2 & 78.5 & 71.1 \\
\cmidrule{2-16}
 & HQQ & 3 & 64 & 30.5 & 6.02 & 3.95 & 80.9 & 52.4 & 63.6 & 77.3 & \textbf{80.7} & 83.9 & \textbf{47.6} & 76.7 & 70.4 \\
 & HQQ & 3 & 128 & 28.5 & 6.48 & 4.35 & 79.8 & 50.4 & 61.6 & 77.0 & \textbf{80.7} & \textbf{85.0} & 45.2 & 76.2 & 69.5 \\
 & EntQuant & 3 & -- & \textbf{25.9} & \textbf{5.74} & \textbf{3.62} & \textbf{81.4} & \textbf{54.0} & \textbf{64.1} & \textbf{78.1} & 80.6 & 84.3 & 47.4 & \textbf{78.4} & \textbf{71.1} \\
\cmidrule{2-16}
 & EntQuant & 2.5 & -- & 22.1 & 6.00 & 3.94 & 80.9 & 50.9 & 62.6 & 77.1 & 80.4 & 82.8 & 46.6 & 78.1 & 69.9 \\
\cmidrule{2-16}
 & HQQ & 2 & 16 & 32.9 & 32.24 & 26.16 & 67.4 & 34.1 & 41.2 & 60.1 & 70.0 & 62.3 & 34.6 & 40.6 & 51.3 \\
 & HQQ & 2 & 32 & 24.9 & 323.93 & 435.23 & 29.8 & 18.4 & 27.6 & 48.9 & 55.7 & 39.9 & 23.4 & 11.1 & 31.9 \\
 & HQQ & 2 & 64 & 20.9 & 2.8e3 & 3.5e3 & 26.9 & 20.6 & 26.3 & 50.7 & 53.4 & 37.8 & 26.4 & 0.8 & 30.4 \\
 & EntQuant & 2.1 & -- & \textbf{18.8} & \textbf{6.47} & \textbf{4.52} & \textbf{78.7} & \textbf{48.8} & \textbf{60.9} & \textbf{74.8} & \textbf{79.9} & \textbf{81.5} & \textbf{43.4} & \textbf{75.4} & \textbf{67.9} \\
\cmidrule{2-16}
 & EntQuant & 1.7 & -- & 15.3 & 8.43 & 6.46 & 71.6 & 40.7 & 54.8 & 68.8 & 76.0 & 69.4 & 39.8 & 60.6 & 60.2 \\
 & EntQuant & 1.3 & -- & 12.7 & 690.33 & 413.31 & 48.9 & 25.2 & 34.9 & 53.0 & 60.3 & 37.9 & 28.6 & 7.2 & 37.0 \\
\bottomrule
\end{tabular}
}
\end{table*}
\begin{table*}[t]
\centering
\caption{Full baseline results for the LLaMA-3.1 base model family (8B, 70B). We report perplexity on C4 and WikiText-2, as well as zero-shot accuracy on eight tasks from LM Eval. All results are generated in-house using the configurations described in \cref{app:implementation}. Best results per model and bit-rate group are in \textbf{bold}. This table extends the results presented in \cref{tab:baseline_main}.}
\label{tab:baseline_full_llama31}
\resizebox{\textwidth}{!}{%
\begin{tabular}{llrrrrrrrrrrrrrr}
\toprule
Model & Method & Bits & Group & Size (GiB) & C4 ↓ & WikiText-2 ↓ & ARC-E ↑ & ARC-C ↑ & HellaSwag ↑ & WinoGrande ↑ & PIQA ↑ & BoolQ ↑ & OBQA ↑ & LAMBADA ↑ & LM Eval Avg. ↑ \\
\midrule
LLaMA-3.1 8B & Base & 16 & -- & 15.0 & 8.43 & 5.84 & 82.0 & 52.0 & 60.8 & 74.3 & 79.4 & 83.0 & 45.4 & 74.8 & 68.9 \\
\cmidrule{2-16}
 & NF4 & 4 & 64 & 5.6 & 8.96 & 6.22 & 81.1 & \textbf{51.5} & 59.5 & 73.5 & 78.7 & \textbf{82.3} & 44.6 & 73.4 & 68.1 \\
 & HQQ & 4 & 64 & 5.6 & 9.02 & 6.26 & 81.2 & 50.9 & 59.7 & \textbf{73.7} & \textbf{79.6} & 81.9 & \textbf{45.4} & \textbf{74.5} & \textbf{68.4} \\
 & HQQ & 4 & 128 & \textbf{5.4} & 9.19 & 6.37 & 80.2 & 48.4 & 59.0 & 73.6 & 78.3 & 81.8 & 44.4 & 73.3 & 67.4 \\
 & EntQuant & 4 & -- & 5.5 & \textbf{8.76} & \textbf{6.08} & \textbf{81.9} & \textbf{51.5} & \textbf{60.1} & 73.3 & 79.1 & \textbf{82.3} & 44.4 & 73.7 & 68.3 \\
\cmidrule{2-16}
 & EntQuant & 3.5 & -- & 5.1 & 9.06 & 6.28 & 80.1 & 49.6 & 59.4 & 72.2 & 78.3 & 82.4 & 44.8 & 73.3 & 67.5 \\
\cmidrule{2-16}
 & HQQ & 3 & 64 & 5.0 & 12.14 & 8.60 & 74.6 & 42.7 & 53.2 & 70.9 & 76.3 & 72.9 & 38.6 & 64.2 & 61.7 \\
 & HQQ & 3 & 128 & 4.8 & 14.46 & 10.31 & 70.1 & 35.8 & 51.1 & 67.5 & 74.2 & 67.2 & 38.6 & 60.8 & 58.2 \\
 & EntQuant & 3 & -- & \textbf{4.7} & \textbf{9.74} & \textbf{6.77} & \textbf{78.2} & \textbf{47.5} & \textbf{57.9} & \textbf{71.7} & \textbf{78.0} & \textbf{80.5} & \textbf{42.6} & \textbf{71.8} & \textbf{66.0} \\
\cmidrule{2-16}
 & EntQuant & 2.6 & -- & 4.4 & 11.25 & 7.87 & 77.6 & 44.3 & 55.5 & 69.4 & 77.8 & 70.7 & 42.6 & 67.5 & 63.2 \\
\cmidrule{2-16}
 & HQQ & 2 & 16 & 5.2 & 142.94 & 163.36 & 33.0 & 19.8 & 28.4 & 50.0 & 57.7 & 52.9 & 24.0 & 2.6 & 33.6 \\
 & HQQ & 2 & 32 & 4.4 & 1.9e3 & 2.9e3 & 25.7 & 19.6 & 26.6 & 52.3 & 53.2 & 37.8 & 24.0 & 0.1 & 29.9 \\
 & HQQ & 2 & 64 & \textbf{4.0} & 1.8e4 & 3.6e4 & 27.0 & 22.1 & 25.7 & 49.6 & 53.6 & 40.1 & 27.2 & 0.0 & 30.7 \\
 & EntQuant & 2.1 & -- & \textbf{4.0} & \textbf{17.56} & \textbf{13.86} & \textbf{60.0} & \textbf{31.2} & \textbf{47.4} & \textbf{62.4} & \textbf{71.1} & \textbf{60.9} & \textbf{37.6} & \textbf{50.1} & \textbf{52.6} \\
\cmidrule{2-16}
 & EntQuant & 1.7 & -- & 3.7 & 135.24 & 186.03 & 42.0 & 21.2 & 34.0 & 53.7 & 61.9 & 42.1 & 27.2 & 12.1 & 36.8 \\
 & EntQuant & 1.4 & -- & 3.4 & 6.4e3 & 3.0e4 & 26.3 & 22.7 & 26.2 & 51.3 & 54.0 & 38.1 & 25.2 & 0.0 & 30.5 \\
\midrule
LLaMA-3.1 70B & Base & 16 & -- & 131.4 & 5.82 & 2.64 & 85.0 & 59.3 & 67.5 & 81.8 & 82.5 & 87.2 & 47.4 & 79.4 & 73.8 \\
\cmidrule{2-16}
 & NF4 & 4 & 64 & 39.8 & 6.35 & 3.07 & 85.8 & 57.9 & 66.7 & 80.7 & \textbf{82.9} & 85.9 & 45.8 & \textbf{78.8} & 73.1 \\
 & HQQ & 4 & 64 & 39.8 & 6.61 & 3.07 & 85.6 & 58.4 & 65.9 & 80.3 & 82.5 & 84.6 & 46.6 & 77.7 & 72.7 \\
 & HQQ & 4 & 128 & 37.8 & 7.08 & 3.47 & 83.3 & 56.1 & 66.0 & 80.6 & 82.1 & 85.3 & \textbf{47.6} & 78.1 & 72.4 \\
 & EntQuant & 3.9 & -- & \textbf{36.2} & \textbf{6.06} & \textbf{2.95} & \textbf{85.9} & \textbf{59.3} & \textbf{67.1} & \textbf{81.1} & 82.6 & \textbf{87.0} & 47.4 & 77.9 & \textbf{73.6} \\
\cmidrule{2-16}
 & EntQuant & 3.5 & -- & 32.9 & 6.26 & 3.21 & 84.8 & 58.3 & 66.6 & 80.7 & 82.2 & 86.4 & 49.0 & 77.9 & 73.2 \\
\cmidrule{2-16}
 & HQQ & 3 & 64 & 33.4 & 629.09 & 327.44 & 54.9 & 22.1 & 29.1 & 54.5 & 64.7 & 47.6 & 31.8 & 7.2 & 39.0 \\
 & HQQ & 3 & 128 & 31.4 & 620.89 & 996.49 & 42.8 & 20.6 & 27.5 & 55.8 & 58.1 & 38.2 & 28.0 & 2.0 & 34.1 \\
 & EntQuant & 3 & -- & \textbf{28.9} & \textbf{6.76} & \textbf{3.76} & \textbf{84.2} & \textbf{58.5} & \textbf{66.0} & \textbf{80.0} & \textbf{81.8} & \textbf{86.4} & \textbf{46.6} & \textbf{77.1} & \textbf{72.6} \\
\cmidrule{2-16}
 & EntQuant & 2.5 & -- & 25.0 & 7.67 & 4.77 & 82.2 & 55.2 & 63.7 & 78.3 & 81.9 & 84.1 & 46.4 & 77.7 & 71.2 \\
\cmidrule{2-16}
 & HQQ & 2 & 16 & 35.8 & 7.7e3 & 8.4e3 & 27.9 & 18.0 & 26.5 & 49.8 & 53.9 & 37.8 & 25.8 & 0.3 & 30.0 \\
 & HQQ & 2 & 32 & 27.8 & 3.0e4 & 3.3e4 & 25.0 & 18.9 & 25.7 & 51.5 & 52.7 & 37.9 & 27.6 & 0.1 & 29.9 \\
 & HQQ & 2 & 64 & 23.8 & 1.3e4 & 2.0e4 & 25.7 & 21.2 & 25.8 & 50.6 & 52.4 & 37.8 & 25.4 & 0.0 & 29.9 \\
 & EntQuant & 2.1 & -- & \textbf{21.7} & \textbf{9.92} & \textbf{6.16} & \textbf{80.6} & \textbf{52.3} & \textbf{61.3} & \textbf{75.7} & \textbf{80.6} & \textbf{80.6} & \textbf{45.8} & \textbf{71.6} & \textbf{68.6} \\
\cmidrule{2-16}
 & EntQuant & 1.7 & -- & 18.2 & 7.9e3 & 1.6e4 & 40.1 & 20.1 & 28.7 & 52.1 & 59.5 & 50.4 & 27.8 & 5.4 & 35.5 \\
 & EntQuant & 1.3 & -- & 15.6 & 2.9e4 & 4.8e4 & 25.4 & 21.9 & 25.8 & 49.5 & 51.6 & 38.3 & 30.0 & 0.0 & 30.3 \\
\bottomrule
\end{tabular}
}
\end{table*}

\clearpage

\section{Full Results for Calibration and Fine-Tuning Methods}
\label{app:results_literature}

\begin{table*}[h!]
\centering
\caption{Comparison of \methodname with calibration and fine-tuning methods on the LLaMA-2 base model family (7B, 13B, 70B). We report perplexity on C4 and WikiText-2, and zero-shot accuracy on five tasks from LM Eval. Results for GPTQ~\citep{frantar2023gptq}, AWQ~\citep{lin2024awq}, OmniQuant~\citep{shao2024omniquant}, LeanQuant~\citep{zhang2025leanquant}, SqueezeLLM~\citep{kim2024squeezellm}, AQLM~\citep{egiazarian2024aqlm}, QuIP\#~\citep{tseng2024quip}, and EfficientQAT~\citep{chen-etal-2025-efficientqat} are taken from~\citet{chen-etal-2025-efficientqat} and \citet{zhang2025leanquant}, with additions from \citet{tseng2024quip} and \citet{shao2024omniquant} for missing perplexity scores. Note that, at 2.1~bits, \methodname operates below the overhead introduced by a group size of 128, which yields 2.14~bits per parameter~\citep[Table~11]{chen-etal-2025-efficientqat}. Best results per model and bit-rate group are in \textbf{bold}. This table extends \cref{tab:literature_main} to include additional model sizes and comparison methods.}
\label{tab:literature_full}
\resizebox{\textwidth}{!}{%
\begin{tabular}{llrrrrrrrrrr}
\toprule
Model & Method & Bits & Group & C4 ↓ & WikiText-2 ↓ & ARC-E ↑ & ARC-C ↑ & HellaSwag ↑ & WinoGrande ↑ & PIQA ↑ & LM Eval Avg. ↑ \\
\midrule
LLaMA-2 7B & Base & 16 & -- & 6.98 & 5.47 & 75.5 & 42.5 & 57.1 & 69.7 & 77.7 & 64.5 \\
\cmidrule{2-12}
 & EntQuant & 3 & -- & 7.55 & 5.95 & 74.6 & \textbf{42.9} & \textbf{56.3} & 68.4 & \textbf{77.7} & \textbf{64.0 (-0.8\%)} \\
 & GPTQ & 3 & 128 & 7.89 & 6.29 & 73.7 & 40.2 & 53.7 & 68.6 & 76.0 & 62.4 (-3.2\%) \\
 & AWQ & 3 & 128 & 7.84 & 6.24 & 74.1 & 41.6 & 55.0 & 67.4 & 76.0 & 62.8 (-2.6\%) \\
 & OmniQuant & 3 & 128 & 7.75 & 6.03 & 74.4 & 39.9 & 54.4 & 66.7 & 76.8 & 62.4 (-3.2\%) \\
 & LeanQuant$_{nu}$ & 3 & -- & 7.73 & 6.19 & 73.7 & 40.2 & 53.2 & 68.3 & 76.4 & 62.4 (-3.3\%) \\
 & SqueezeLLM & 3 & -- & 7.72 & 6.18 & 73.1 & 40.3 & 54.1 & 67.9 & 76.5 & 62.4 (-3.3\%) \\
 & QuIP\# & 3 & -- & \textbf{7.32} & \textbf{5.79} & 74.6 & 41.9 & 55.9 & 68.2 & 77.0 & 63.5 (-1.5\%) \\
 & EfficientQAT & 3 & 128 & 7.34 & 5.81 & \textbf{74.7} & 42.8 & 55.9 & \textbf{69.1} & 77.6 & \textbf{64.0 (-0.8\%)} \\
\cmidrule{2-12}
 & EntQuant & 2.1 & -- & 10.59 & 8.29 & 68.9 & 36.2 & 51.6 & 65.7 & 73.7 & 59.2 (-8.3\%) \\
 & GPTQ & 2 & 128 & 33.70 & 36.77 & 40.5 & 21.2 & 32.6 & 55.2 & 58.3 & 41.6 (-35.6\%) \\
 & OmniQuant & 2 & 128 & 15.02 & 11.06 & 50.1 & 23.5 & 40.3 & 55.9 & 65.1 & 47.0 (-27.2\%) \\
 & LeanQuant$_{nu}$ & 2 & -- & 17.07 & 15.51 & 51.8 & 24.0 & 35.9 & 58.2 & 66.4 & 47.2 (-26.8\%) \\
 & AQLM & 2 & 1x16 & -- & -- & \textbf{74.1} & \textbf{39.7} & \textbf{53.4} & 65.2 & \textbf{76.9} & \textbf{61.8 (-4.1\%)} \\
 & QuIP\# & 2 & -- & \textbf{8.35} & \textbf{6.66} & 71.8 & 37.9 & 52.2 & 65.7 & 75.5 & 60.6 (-6.1\%) \\
 & EfficientQAT & 2 & 64 & 8.50 & 6.86 & 71.0 & 36.9 & 51.6 & 66.0 & 75.3 & 60.1 (-6.8\%) \\
 & EfficientQAT & 2 & 128 & 8.79 & 7.19 & 69.8 & 36.5 & 50.8 & \textbf{66.2} & 74.2 & 59.5 (-7.8\%) \\
\midrule
LLaMA-2 13B & Base & 16 & -- & 6.47 & 4.88 & 78.9 & 47.2 & 60.2 & 72.2 & 79.7 & 67.6 \\
\cmidrule{2-12}
 & EntQuant & 3 & -- & 6.76 & 5.15 & 77.3 & 44.3 & \textbf{59.4} & 72.1 & 78.5 & 66.3 (-2.0\%) \\
 & GPTQ & 3 & 128 & 7.00 & 5.42 & 78.0 & 45.6 & 57.8 & 70.9 & \textbf{78.6} & 66.2 (-2.1\%) \\
 & AWQ & 3 & 128 & 6.94 & 5.32 & 78.0 & 44.6 & 58.6 & 71.8 & 77.8 & 66.1 (-2.2\%) \\
 & OmniQuant & 3 & 128 & 6.98 & 5.28 & 77.9 & 46.2 & 58.5 & 70.0 & 78.4 & 66.2 (-2.2\%) \\
 & LeanQuant$_{nu}$ & 3 & -- & 6.98 & 5.40 & 77.2 & 44.2 & 56.4 & 70.1 & 77.8 & 65.1 (-3.7\%) \\
 & SqueezeLLM & 3 & -- & 6.97 & 5.36 & 77.3 & 43.2 & 58.7 & 69.5 & 77.9 & 65.3 (-3.4\%) \\
 & QuIP\# & 3 & -- & \textbf{6.72} & \textbf{5.10} & 77.9 & 44.6 & 58.3 & \textbf{72.5} & 78.1 & 66.3 (-2.0\%) \\
 & EfficientQAT & 3 & 128 & 6.73 & 5.12 & \textbf{79.0} & \textbf{48.0} & 59.0 & 72.1 & 78.4 & \textbf{67.3 (-0.5\%)} \\
\cmidrule{2-12}
 & EntQuant & 2.1 & -- & 7.95 & 6.24 & 74.5 & 41.4 & 55.6 & 69.6 & 76.9 & 63.6 (-6.0\%) \\
 & GPTQ & 2 & 128 & 20.97 & 28.14 & 55.6 & 21.9 & 41.1 & 55.8 & 67.1 & 48.3 (-28.6\%) \\
 & OmniQuant & 2 & 128 & 11.05 & 8.26 & 63.2 & 30.3 & 46.2 & 57.9 & 70.1 & 53.6 (-20.8\%) \\
 & LeanQuant$_{nu}$ & 2 & -- & 11.83 & 10.06 & 62.5 & 30.2 & 42.2 & 62.0 & 69.9 & 53.4 (-21.1\%) \\
 & AQLM & 2 & 1x16 & -- & -- & 75.2 & \textbf{43.5} & \textbf{57.6} & \textbf{70.1} & \textbf{78.3} & \textbf{65.0 (-4.0\%)} \\
 & QuIP\# & 2 & -- & \textbf{7.45} & \textbf{5.74} & \textbf{75.7} & 42.9 & 56.5 & 69.1 & 78.0 & 64.4 (-4.7\%) \\
 & EfficientQAT & 2 & 64 & 7.59 & 5.96 & 74.8 & 41.9 & 55.3 & 68.4 & 77.0 & 63.5 (-6.1\%) \\
 & EfficientQAT & 2 & 128 & 7.75 & 6.08 & 75.0 & 42.8 & 55.7 & 68.9 & 77.0 & 63.9 (-5.5\%) \\
\midrule
LLaMA-2 70B & Base & 16 & -- & 5.52 & 3.32 & 82.6 & 54.4 & 65.3 & 80.2 & 81.6 & 72.8 \\
\cmidrule{2-12}
 & EntQuant & 3 & -- & 5.74 & 3.62 & 81.4 & 54.0 & 64.1 & \textbf{78.1} & 80.6 & 71.7 (-1.6\%) \\
 & GPTQ & 3 & 128 & 5.85 & 3.85 & 81.7 & 53.7 & 62.9 & 77.7 & 81.5 & 71.5 (-1.9\%) \\
 & AWQ & 3 & 128 & 5.81 & 3.74 & 81.4 & 53.7 & 63.7 & 76.5 & 81.8 & 71.4 (-1.9\%) \\
 & OmniQuant & 3 & 128 & 5.85 & 3.78 & 81.0 & 52.8 & 63.5 & 76.5 & 81.5 & 71.1 (-2.4\%) \\
 & QuIP\# & 3 & -- & \textbf{5.67} & \textbf{3.56} & \textbf{82.1} & \textbf{55.9} & \textbf{64.2} & 76.2 & \textbf{82.2} & \textbf{72.1 (-0.9\%)} \\
 & EfficientQAT & 3 & 128 & 5.71 & 3.61 & 81.7 & 53.8 & \textbf{64.2} & 77.3 & 81.8 & 71.8 (-1.5\%) \\
\cmidrule{2-12}
 & EntQuant & 2.1 & -- & 6.47 & 4.52 & 78.7 & 48.8 & 60.9 & 74.8 & 79.9 & 68.6 (-5.8\%) \\
 & GPTQ & 2 & 128 & -- & -- & 25.1 & 22.7 & 25.0 & 49.6 & 49.5 & 34.4 (-52.8\%) \\
 & OmniQuant & 2 & 128 & 8.52 & 6.55 & 67.2 & 33.3 & 35.5 & 64.3 & 74.1 & 54.9 (-24.6\%) \\
 & AQLM & 2 & 1x16 & -- & -- & 81.4 & \textbf{53.0} & 62.8 & \textbf{76.0} & 81.1 & 70.8 (-2.7\%) \\
 & QuIP\# & 2 & -- & \textbf{6.12} & \textbf{4.16} & \textbf{81.9} & 52.6 & \textbf{62.9} & 75.8 & \textbf{81.4} & \textbf{70.9 (-2.6\%)} \\
 & EfficientQAT & 2 & 64 & 6.38 & 4.52 & 80.1 & 50.8 & 61.8 & 74.6 & 80.1 & 69.5 (-4.6\%) \\
 & EfficientQAT & 2 & 128 & 6.48 & 4.61 & 80.0 & 49.2 & 61.6 & 73.6 & 80.2 & 68.9 (-5.3\%) \\
\bottomrule
\end{tabular}
}
\end{table*}

\clearpage

\section{Full Results for Instruction-Tuned Models}
\label{app:results_instruct}

\begin{table*}[h!]
\centering
\caption{Full results on a variety of instruction-tuned models across four advanced benchmarks from LM Eval: instruction-following (IFEval), mathematical reasoning (GSM8K with chain-of-thought), scientific reasoning (GPQA), and broad knowledge (MMLU); see \cref{app:implementation:eval} for more details on the benchmark configurations. These results extend the visualization of \cref{fig:figure1}.}
\label{tab:instruct_full}
\resizebox{\textwidth}{!}{%
\begin{tabular}{llrrrrrrrr}
\toprule
Model & Method & Bits & C4 ↓ & WikiText-2 ↓ & GSM8K ↑ & GPQA ↑ & MMLU ↑ & IFEval ↑ & LM Eval Avg. ↑ \\
\midrule
LLaMA-3.1 8B Instruct & Base & 16 & 9.75 & 6.76 & 84.7 & 30.8 & 68.7 & 73.6 & 64.4 \\
 & EntQuant & 4 & 10.11 & 7.02 & 83.7 & 29.9 & 67.5 & 76.0 & 64.3 \\
 & EntQuant & 3 & 11.11 & 7.73 & 75.1 & 28.6 & 63.9 & 70.8 & 59.6 \\
 & EntQuant & 2.2 & 17.95 & 13.57 & 14.6 & 24.6 & 48.6 & 49.0 & 34.2 \\
\midrule
LLaMA-3.1 70B Instruct & Base & 16 & 6.74 & 3.59 & 94.6 & 41.5 & 83.1 & 84.5 & 75.9 \\
 & EntQuant & 3.9 & 7.04 & 3.83 & 94.5 & 42.0 & 82.6 & 85.0 & 76.0 \\
 & EntQuant & 3 & 7.72 & 4.51 & 93.3 & 37.9 & 81.3 & 82.6 & 73.8 \\
 & EntQuant & 2.1 & 11.23 & 6.76 & 87.4 & 37.7 & 75.6 & 78.0 & 69.7 \\
\midrule
LLaMA-3.3 70B Instruct & Base & 16 & 6.87 & 3.62 & 94.3 & 50.4 & 82.0 & 88.9 & 78.9 \\
 & EntQuant & 3.9 & 7.19 & 3.94 & 93.9 & 50.9 & 81.9 & 87.6 & 78.6 \\
 & EntQuant & 3 & 7.89 & 4.63 & 93.7 & 44.2 & 80.6 & 87.6 & 76.5 \\
 & EntQuant & 2.1 & 11.29 & 6.97 & 86.7 & 35.7 & 75.7 & 86.1 & 71.1 \\
\midrule
Qwen3 8B & Base & 16 & 13.30 & 9.73 & 81.4 & 39.3 & 72.0 & 81.7 & 68.6 \\
 & EntQuant & 3.9 & 13.62 & 9.96 & 76.0 & 37.5 & 69.6 & 82.3 & 66.3 \\
 & EntQuant & 3 & 14.76 & 11.24 & 75.4 & 34.4 & 63.3 & 78.7 & 63.0 \\
 & EntQuant & 2.2 & 20.12 & 17.64 & 28.7 & 23.2 & 30.3 & 56.9 & 34.8 \\
\midrule
Qwen3 14B & Base & 16 & 12.02 & 8.64 & 82.8 & 39.3 & 76.7 & 84.5 & 70.8 \\
 & EntQuant & 3.9 & 12.29 & 8.87 & 80.2 & 37.5 & 76.5 & 84.3 & 69.6 \\
 & EntQuant & 3 & 12.70 & 9.29 & 82.8 & 31.5 & 74.3 & 84.3 & 68.2 \\
 & EntQuant & 2.2 & 14.98 & 11.61 & 68.0 & 28.8 & 57.9 & 76.3 & 57.8 \\
\midrule
Qwen3 32B & Base & 16 & 10.78 & 7.61 & 89.2 & 42.6 & 80.9 & 84.3 & 74.2 \\
 & EntQuant & 3.9 & 10.98 & 7.79 & 86.3 & 43.3 & 80.4 & 82.8 & 73.2 \\
 & EntQuant & 3 & 11.32 & 8.13 & 84.7 & 39.7 & 78.2 & 80.8 & 70.9 \\
 & EntQuant & 2.2 & 13.47 & 10.45 & 79.9 & 37.3 & 71.7 & 78.7 & 66.9 \\
\midrule
OLMo-3.1 32B Instruct & Base & 16 & 11.52 & 6.95 & 93.9 & 37.7 & 74.1 & 87.4 & 73.3 \\
 & EntQuant & 3.9 & 11.59 & 7.05 & 93.7 & 37.5 & 73.8 & 87.2 & 73.1 \\
 & EntQuant & 3 & 11.66 & 7.21 & 93.3 & 38.8 & 73.3 & 88.0 & 73.4 \\
 & EntQuant & 2.1 & 12.71 & 8.43 & 93.4 & 35.5 & 70.4 & 84.3 & 70.9 \\
\midrule
Mistral Large Instruct 24.11 123B & Base & 16 & 5.53 & 2.64 & 84.2 & 47.5 & 83.1 & 81.9 & 74.2 \\
 & EntQuant & 3.9 & 5.60 & 2.73 & 85.4 & 45.8 & 82.9 & 81.7 & 73.9 \\
 & EntQuant & 3 & 5.80 & 2.99 & 84.7 & 46.2 & 82.4 & 79.3 & 73.1 \\
 & EntQuant & 2.1 & 6.63 & 4.01 & 82.6 & 42.9 & 76.8 & 78.7 & 70.3 \\
\bottomrule
\end{tabular}
}
\end{table*}

\clearpage

\section{Full Results for Model Inference}
\label{app:results_inference}

\begin{figure}[h!]
    \centering
    \includegraphics[width=\linewidth]{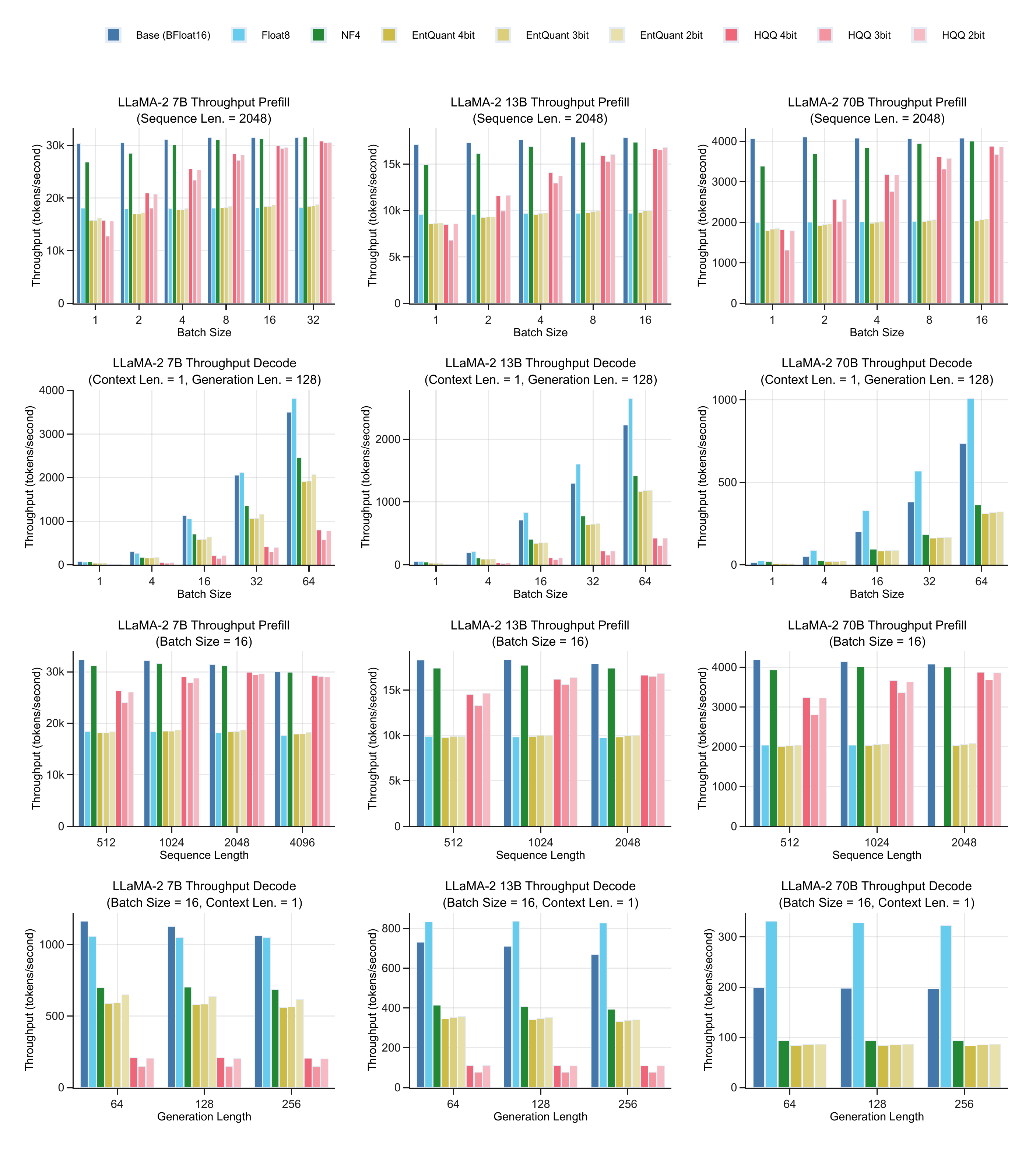}
    \caption{Inference \textbf{throughput} comparison across the LLaMA-2 model base family (7B, 13B, 70B) under varying inference configurations. Top two rows: Prefill and decode throughput as a function of batch size. Bottom two rows: Prefill and decode throughput as a function of sequence/generation length. Missing values are due to CUDA out-of-memory errors and/or exceptionally long runtimes. \methodnameplain benefits from the Marlin kernel's superior performance, particularly in pure decoding regimes (input context length 1). In prefill-dominated regimes, both \floateightplain and \methodnameplain are slightly slower than the \bfloatplain baseline due to kernel overhead. These results extend \cref{fig:efficiency_main}.}
    \label{fig:efficiency_full_throughput}
\end{figure}

\begin{figure}[t]
    \centering
    \includegraphics[width=\linewidth]{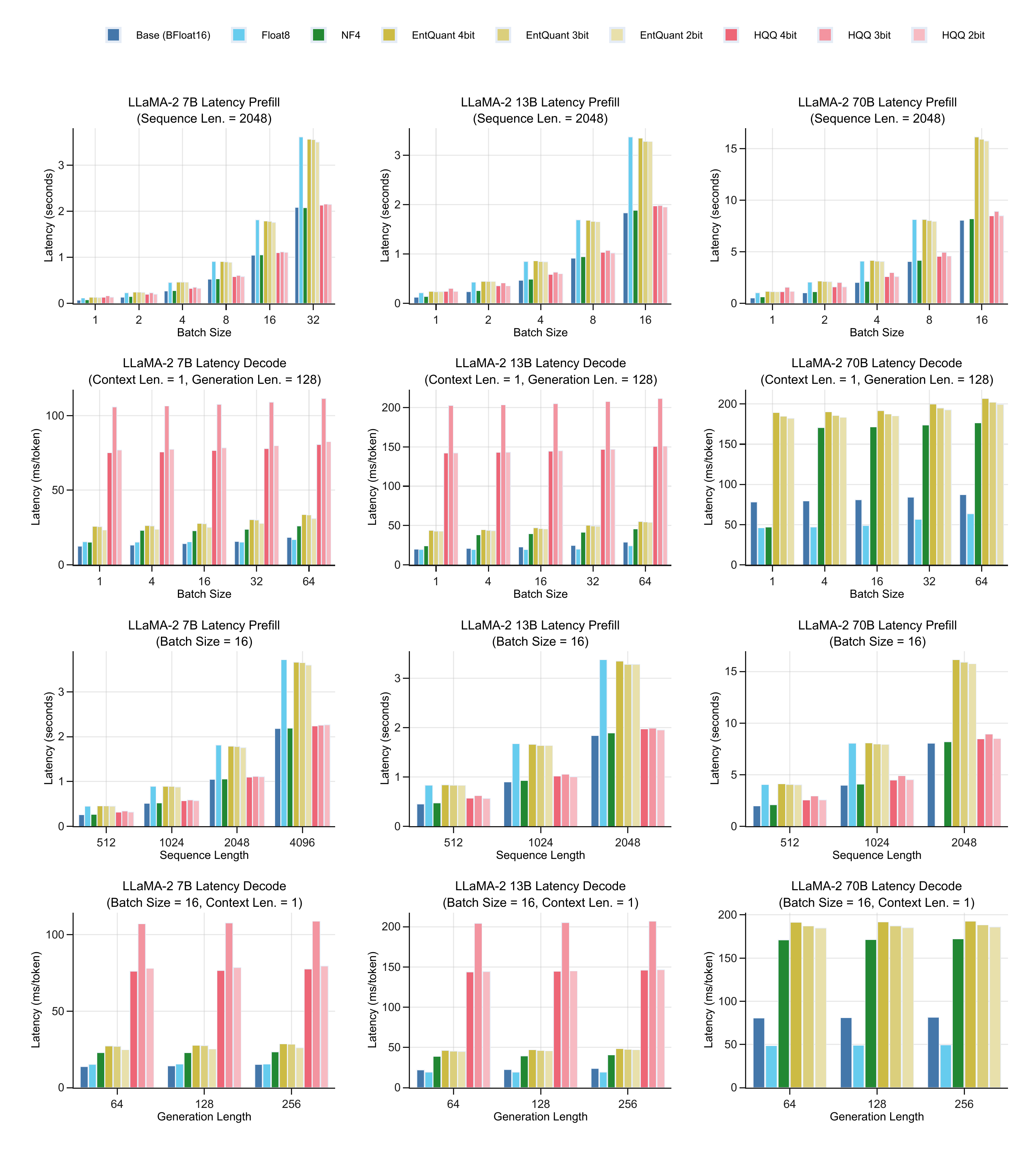}
    \caption{Inference \textbf{latency} comparison across the LLaMA-2 model base family (7B, 13B, 70B) under varying inference configurations. Top two rows: Prefill and decode latency as a function of batch size. Bottom two rows: Prefill and decode latency as a function of sequence/generation length. Missing values are due to CUDA out-of-memory errors and/or exceptionally long runtimes. These results extend \cref{fig:efficiency_main}.}
    \label{fig:efficiency_full_latency}
\end{figure}

\begin{figure}[t]
    \centering
    \includegraphics[width=\linewidth]{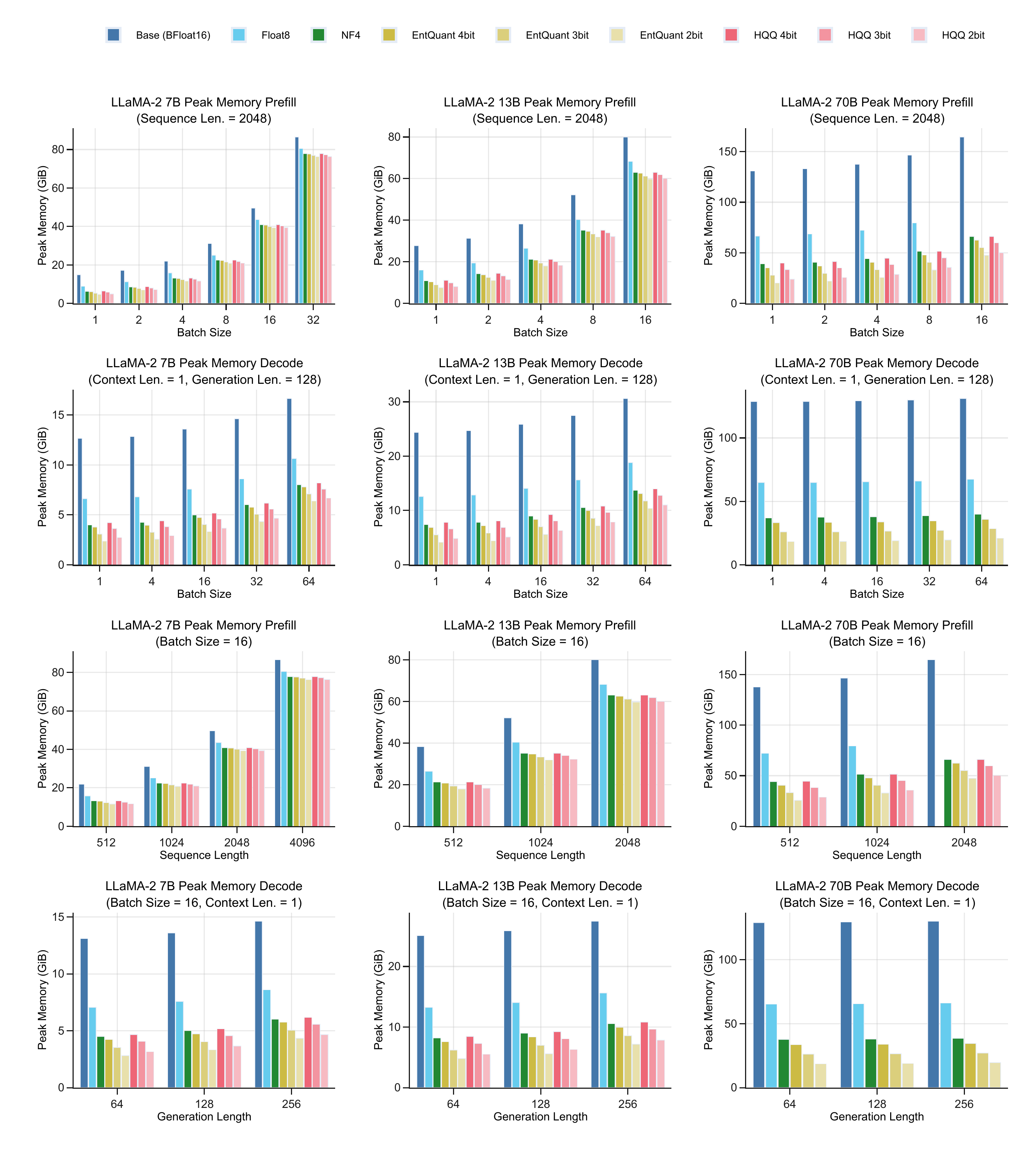}
    \caption{Inference \textbf{peak memory} comparison across the LLaMA-2 model base family (7B, 13B, 70B) under varying inference configurations. Top two rows: Prefill and decode peak memory as a function of batch size. Bottom two rows: Prefill and decode peak memory as a function of sequence/generation length. Missing values are due to CUDA out-of-memory errors and/or exceptionally long runtimes. These results extend \cref{fig:efficiency_main}.}
    \label{fig:efficiency_full_peak_memory}
\end{figure}

\clearpage

\section{Full Results for \methodnameplain Float8 vs.~Int8 and Super Weight Handling}
\label{app:results_sw_ablation}

\begin{table*}[h!]
\centering
\caption{Comparison of \methodnameplain with \floateightplain and \inteightplain, including an ablation study on super weight handling for the LLaMA-2 base model family (7B, 13B, 70B). We compare \methodnameplain with \floateightplain and \inteightplain base formats under different super weight (SW) exclusion thresholds: 10, 50, and $\infty$ (no exclusion). Lower thresholds declare more weights as super weights, leading to more excluded layers; see \cref{sec:experiments:sw} for more details. For comparison, we also show NF4 and HQQ results with and without super weight exclusion (using threshold 50, which is the default in \citet{yu2025superweightlargelanguage}). These results extend the analysis of \cref{fig:sw_ablation}.}
\label{tab:sw_ablation}
\begin{tabular}{lrrrrrrrr}
\toprule
 &  &  & \multicolumn{3}{c}{C4 ↓ (Perplexity)} & \multicolumn{3}{c}{WikiText-2 ↓ (Perplexity)} \\
\cmidrule(lr){4-6}
\cmidrule(lr){7-9}
Method & Bits & Group & 2-7 & 2-13 & 2-70 & 2-7 & 2-13 & 2-70 \\
\midrule
Base & 16 & -- & 6.98 & 6.47 & 5.52 & 5.47 & 4.88 & 3.32 \\
\midrule
NF4 & 4 & 64 & 7.16 & 6.57 & 5.59 & 5.65 & 4.98 & 3.42 \\
NF4 (SW) & 4 & 64 & 7.10 & 6.55 & 5.57 & 5.59 & 4.97 & 3.40 \\
HQQ & 4 & 64 & 7.20 & 6.57 & 5.60 & 5.68 & 4.98 & 3.43 \\
HQQ (SW) & 4 & 64 & 7.10 & 6.55 & 5.57 & 5.60 & 4.96 & 3.39 \\
EntQuant Float8 (SW 10) & 4 & -- & 7.07 & 6.53 & 5.59 & 5.56 & 4.95 & 3.39 \\
EntQuant Float8 (SW 50) & 3.9/4 & -- & 7.07 & 6.53 & 5.59 & 5.57 & 4.96 & 3.39 \\
EntQuant Float8 (SW Inf) & 3.9 & -- & 7.14 & 6.55 & 5.59 & 5.62 & 4.97 & 3.40 \\
EntQuant Int8 (SW 10) & 4 & -- & 7.07 & 6.53 & 5.57 & 5.56 & 4.96 & 3.38 \\
EntQuant Int8 (SW 50) & 3.9/4 & -- & 7.08 & 11.54 & 5.57 & 5.56 & 7.82 & 3.39 \\
EntQuant Int8 (SW Inf) & 3.9 & -- & 21.42 & 13.13 & 5.97 & 22.10 & 9.55 & 3.83 \\
\midrule
HQQ & 3 & 64 & 9.11 & 7.29 & 6.02 & 7.05 & 5.60 & 3.95 \\
HQQ (SW) & 3 & 64 & 7.76 & 6.93 & 5.86 & 6.16 & 5.33 & 3.77 \\
EntQuant Float8 (SW 10) & 3.1 & -- & 7.41 & 6.72 & 5.71 & 5.86 & 5.12 & 3.57 \\
EntQuant Float8 (SW 50) & 3/3.1 & -- & 7.45 & 6.73 & 5.71 & 5.87 & 5.12 & 3.57 \\
EntQuant Float8 (SW Inf) & 3 & -- & 7.55 & 6.76 & 5.74 & 5.95 & 5.15 & 3.62 \\
EntQuant Int8 (SW 10) & 3.1 & -- & 7.41 & 6.72 & 5.68 & 5.86 & 5.12 & 3.55 \\
EntQuant Int8 (SW 50) & 3/3.1 & -- & 7.45 & 6.73 & 5.68 & 5.87 & 5.14 & 3.56 \\
EntQuant Int8 (SW Inf) & 3 & -- & 7.61 & 6.75 & 5.76 & 5.99 & 5.16 & 3.63 \\
\midrule
HQQ & 2 & 64 & 6.0e3 & 2.7e3 & 2.8e3 & 7.5e3 & 3.8e3 & 3.5e3 \\
HQQ (SW) & 2 & 64 & 910.97 & 357.99 & 2.4e3 & 1.2e3 & 395.38 & 3.0e3 \\
EntQuant Float8 (SW 10) & 2.2 & -- & 9.75 & 7.98 & 6.34 & 7.84 & 6.26 & 4.44 \\
EntQuant Float8 (SW 50) & 2.1 & -- & 10.67 & 8.05 & 6.35 & 8.44 & 6.36 & 4.46 \\
EntQuant Float8 (SW Inf) & 2 & -- & 11.33 & 8.10 & 6.59 & 8.84 & 6.39 & 4.66 \\
EntQuant Int8 (SW 10) & 2.2 & -- & 9.74 & 7.96 & 6.33 & 7.82 & 6.24 & 4.43 \\
EntQuant Int8 (SW 50) & 2.1 & -- & 10.65 & 8.03 & 6.34 & 8.42 & 6.33 & 4.44 \\
EntQuant Int8 (SW Inf) & 2 & -- & 11.31 & 8.09 & 6.55 & 8.81 & 6.36 & 4.62 \\
\bottomrule
\end{tabular}
\end{table*}

\clearpage

\end{document}